# Human-AI Governance (HAIG): A Trust-Utility Approach

Zeynep Engin

The Digital Statecraft Academy, United Kingdom
Department of Computer Science, University College London (UCL), United Kingdom

Correspondence: z.engin@digitalstatecraft.academy

## Abstract

This paper introduces the Human-AI Governance (HAIG) framework, contributing to the AI Governance (AIG) field by foregrounding the relational dynamics between human and AI actors rather than treating AI systems as objects of governance alone. Current categorical frameworks (e.g., 'human-in-the-loop' models) inadequately capture how AI systems evolve from 'tools' to 'partners', particularly as foundation models demonstrate emergent capabilities and multi-agent systems exhibit autonomous goal-setting behaviours. As systems are deployed across contexts, agency redistributes in complex patterns that are better represented as positions along continua rather than discrete categories. The HAIG framework operates across three levels: *dimensions* (Decision Authority, Process Autonomy, and Accountability Configuration), *continua* (continuous positional spectra along each dimension), and *thresholds* (critical points along the continua where governance requirements shift qualitatively). The framework's dimensional architecture is level-agnostic, applicable from individual deployment decisions and organisational governance through to sectorial comparison and national/international regulatory design. Unlike risk-based or principle-based approaches that treat governance primarily as a constraint on AI deployment, HAIG adopts a *trust-utility* orientation – reframing governance as the condition under which human-AI collaboration can realise its potential, calibrating oversight to specific relational contexts rather than predetermined categories. The analysis reveals how technical advances in self-supervision, reasoning authority, and distributed decision-making drive context-dependent and non-uniform evolution in trustworthiness conditions — with consequences for the trust attitudes of relevant stakeholders that existing categorical frameworks are ill-equipped to track. Case studies in healthcare and European regulation demonstrate how HAIG complements existing frameworks while offering a foundation for adaptive regulatory design that anticipates governance challenges before they emerge.

# 1. Introduction: Rethinking Human-AI Relationships

The governance of AI systems increasingly demands frameworks capable of accounting for dynamic, non-uniform shifts in how authority, autonomy, and accountability are distributed between human and AI actors. As algorithmic systems take on greater roles in shaping decisions, allocating resources, and influencing public life, the question is not merely whether we trust them, but whether they are designed and maintained to warrant that trust. What is required is neither blind faith in machines nor paralysing scepticism, but a governance model that can continually calibrate *how much trust is appropriate, and in which contexts*.

Artificial intelligence is fundamentally reshaping the landscape of organisational decision-making, public service delivery, and institutional governance (Engin et al., 2025; Rudko et al., 2024; Shrestha et al., 2019). Across sectors, it is rapidly evolving from basic automation to increasingly sophisticated forms of governance partnership (Rahwan et al., 2019; Pedreschi et al., 2025; Liu et al., 2025), aiming for human–AI complementarity where collaboration produces greater value than either alone (Stephany & Teutloff, 2024; Akata et al., 2020; Donahue et al., 2022; van Breda & Barry, 2020). This evolution is particularly evident in technical advances such as foundation models with emergent capabilities, self-supervised learning systems that critique their own outputs, and multi-agent architectures that exhibit collective behaviours beyond their individual components. These technical developments fundamentally challenge traditional governance frameworks by introducing dynamic capability boundaries and distributed agency patterns that resist static categorisation. However, our conceptual frameworks have not kept pace with this evolution (Goos & Savona, 2024).

This limitation becomes particularly evident as AI systems increasingly permeate domains traditionally dominated by human judgment. In healthcare, diagnostic AI systems gradually shift from flagging anomalies to making preliminary diagnoses (Yelne et al., 2023; Qiu et al., 2024); in financial services, credit algorithms evolve from augmenting human judgment to making autonomous decisions for standard applications (Gsenger & Strle, 2021); in transportation, self-driving systems dynamically redistribute authority between human and machine based on environmental conditions (Ansari et al., 2022). Even in knowledge work requiring complex reasoning—legal practice, academic research, strategic decision-making, and creative production—AI systems now operate as collaborative partners rather than mere tools (Sowa & Przegalinska, 2025; Koehler & Sauermann, 2024), progressively assuming greater agency. Multi-agent systems embedded in infrastructure further demonstrate this evolution, with interconnected algorithms collectively reshaping environments through distributed decision-making (Cugurullo & Xu, 2025; Luusua et al., 2023; Engin et al., 2020). Perhaps most consequentially, AI is increasingly embedded in processes central to democratic governance —from content moderation that shapes public discourse, to algorithmic allocation of public resources, to automated enforcement of regulations, to decision support systems that inform policy development (Engin & Treleaven, 2019; Dunleavy & Margetts, 2023). AI systems in these contexts fundamentally challenge traditional frameworks for understanding accountability, oversight, and democratic control (Coglianese & Lehr, 2017; Gritsenko & Wood, 2020; Stokel, 2025) as they incrementally assume greater decision-making authority without corresponding evolutions in oversight frameworks.

Current approaches to understanding human-AI relationships rely on discrete categories—tools versus agents, decision support versus automated decision-making, or human-in-the-loop versus human-out-of-the-loop models. These categorical frameworks provide implementation clarity but fail to account for the hybrid, evolutionary nature of AI systems in practice (Engin & Hand, 2025; Coglianese & Crum, 2025; Earp et al., 2025; Gabriel et al., 2024).

This paper introduces the Human-AI Governance (HAIG) framework for analysing *trust dynamics* across evolving human-AI relationships. Situated within but extending the emerging field of AI governance (AIG), which Birkstedt et al. (2023) characterise as spanning ethical, regulatory, organisational, and technical domains, HAIG foregrounds the relational dimension: how *decision authority*, *process autonomy*, and *accountability configuration* are dynamically distributed between human and AI actors across evolving deployment contexts. Where existing AIG frameworks are predominantly harm-preventive — designed to identify risks, assign categories, and impose constraints on AI deployment (Kaminski, 2023; Ebers, 2024) — HAIG's *trust-utility* orientation reframes this: governance is not only a constraint on AI capability but a condition for realising its potential. Utility here encompasses not only technical effectiveness but democratic legitimacy and public confidence; performance gains achieved at the cost of accountability diffusion do not satisfy the *trust-utility* criterion. By calibrating oversight to the trust requirements of specific human-AI relationships rather than to predetermined risk categories, HAIG makes proactive governance tractable — anticipating where existing mechanisms will become insufficient before capabilities advance past them, rather than responding reactively once governance gaps have become acute. HAIG operates across three complementary levels: *dimensions*, *continua*, and *thresholds*, providing both theoretical comprehensiveness and practical flexibility by acknowledging continuous agency redistribution while identifying meaningful transition points requiring governance adaptation. Throughout, *human-AI relationship* refers to the governance relationship between human and AI actors — how authority, autonomy, and accountability are distributed across a given deployment context — rather than to the interactional or psychological dimensions of human-AI engagement.

The remainder of this paper unfolds as follows: Section 2 sets out the research approach and nature of the contribution; Section 3 examines current approaches and their limitations, establishing the structural governance deficit that motivates the framework; Section 4 introduces the HAIG framework — the paper's primary original contribution — elaborating its three dimensions, continuum architecture, and trust threshold system; Section 5 demonstrates practical applications through healthcare and EU regulatory case studies; and Section 6 discusses broader implications and future research directions.

# 2. Research Approach and Nature of Contribution

The HAIG framework is a conceptual contribution in the tradition of integrative theory development (Jaakkola, 2020): it synthesises existing bodies of knowledge into a new analytical architecture rather than generating new empirical data. The governance gap it addresses — a structural mismatch between the continuous, multi-dimensional nature of human-AI relationships and the static categorical frameworks developed to govern them — motivates a dimensional approach in which governance relationships are represented as positions along continuous axes rather than as members of discrete categories. That gap has two structural foundations that any adequate response must address (Narayanan and Kapoor, 2024). First, risk and governance requirements are not properties of AI models but emerge from deployment context: the same system may present negligible governance concerns in one application and require stringent oversight in another. Second, unlike traditional regulatory objects such as pharmaceuticals or aircraft, AI systems are not static artefacts — they are continuously updated, fine-tuned, and version-dependent, such that a system approved or categorised under one configuration may behave substantially differently after subsequent modification. Static categorical frameworks, by design, cannot track either form of variation systematically. Dimensional positioning can.

Framework development proceeded through structured synthesis of three bodies of scholarship. Human-automation interaction theory (Parasuraman et al., 2000; Vagia et al., 2016; Methnani et al., 2021) provides the theoretical vocabulary for continuous representation of authority and autonomy. AI governance and regulatory scholarship (Mäntymäki et al., 2022; Birkstedt et al., 2023; Kaminski, 2023; Smuha, 2021) establishes the governance gap the framework addresses and situates HAIG primarily within the organisational and regulatory levels of AIG analysis. Trust theory as applied to automated systems (Lee & See, 2004; Toreini et al., 2020; Ferrario et al. 2024; Laux et al., 2024) grounds the *trust-utility* orientation and the threshold architecture. These three clusters have largely developed in parallel; their integration is itself a substantive contribution. The dimensional approach it produces is not novel in principle — Wilson et al. (2025) demonstrate its analytical value within clinical decision support — but HAIG applies it at a different level of abstraction: a cross-sector governance architecture built on three theoretically orthogonal axes of Authority, Autonomy, and Accountability (Engin & Hand, 2025).

By design, HAIG's dimensional architecture is level-agnostic: the same continua apply whether calibrating oversight for a single deployment, comparing governance positioning across sectors, or evaluating the implications of an emerging regulatory paradigm. The organisational and institutional level (Mäntymäki et al., 2022) is the primary locus of directly actionable outputs in this paper — reflected in the framework's deployment guidance (Section 4.4) and the healthcare case study (Section 5.1) — while the sectorial and national regulatory levels are addressed through Figure 2 and the EU AI Act analysis (Sections 5.2–5.3). The individual and international levels represent natural extensions of the framework's logic, and are identified as future research directions in Section 6.

HAIG's most significant limitation is the absence of longitudinal empirical validation tracking how governance frameworks evolve alongside AI capabilities across real deployments — the primary future research direction identified in Section 6. A related limitation concerns operationalisation: authority, autonomy, and accountability do not correspond to simple physical measurements, and developing robust validated measures for them is identified as important future work. The trust thresholds in Section 4.3 are designed to generate testable propositions toward this end, making specific predictions about where governance frameworks will encounter difficulty that can be evaluated against empirical evidence as it accumulates.

To guard against the risk that disciplinary integration coherent in the abstract may fail to capture the practical texture of the phenomena it describes, the framework is grounded in cross-sector governance practice — spanning applied research collaborations in healthcare, public administration, and regulatory contexts; sustained editorial engagement with interdisciplinary scholarship at the intersection of data science and public policy; and leadership roles within international AI governance networks. The recurring patterns this practice surfaces — static categorical frameworks that break down at boundaries, trust mechanisms that fail to scale, accountability structures that cannot accommodate distributed agency — are not theoretical constructs but observed governance failures that motivate the framework's design, consistent with design science approaches to framework development (Hevner et al., 2004).

# 3. Governing Human-AI Relationships: Current Approaches and Their Limitations

AI integration across public and commercial domains demands new approaches to human-AI relationships (Lai et al., 2023; Rahwan, 2018; Wilson et al., 2025). This section maps the landscape of existing approaches across four strands that collectively motivate the HAIG framework — not arbitrarily, but because their limitations, taken together, define the governance gap it addresses. Governance philosophy scholarship (Section 3.1) establishes what existing regulatory approaches can and cannot do. Trust and accountability studies (Section 3.2) reveal how confidence in algorithmic systems is built and eroded over time. Agentic AI scholarship (Section 3.3) documents the technical developments outpacing those frameworks. Section 3.4 then synthesises these limitations.

## 3.1. Governance Philosophies

As AI systems increasingly permeate decision-making contexts, several distinct governance approaches have emerged to address the challenges of human-AI relationships. These approaches reflect different priorities, disciplinary perspectives, and governance philosophies that shape how we conceptualise, implement, and regulate AI systems across society.

*Risk-based* approaches exemplify the tension between clarity and adaptability. The EU's AI Act categorises systems by potential harm (unacceptable to minimal risk), yet struggles with evolving systems and contextual variation (Kaminski, 2023; Laux et al., 2024). Foundation models particularly challenge this approach with their 'multifunctionality' across deployment contexts (Coglianese & Crum, 2025). *Rights-based* frameworks centre on protecting fundamental human rights in AI deployment (Smuha, 2021; UN Human Rights Council, 2024; Toronto Declaration, 2018). While establishing essential normative foundations, they often lack specific implementation guidance for complex socio-technical systems. *Principle-based* governance, advanced by organisations like the OECD and IEEE, establishes broad normative guidelines focusing on values like fairness, transparency, and accountability. While creating useful ethical frameworks, these approaches often struggle with operationalisation and enforcement in complex technical contexts (Mittelstadt, 2019). *Trust-centred* approaches position trust dynamics as fundamental to effective human-AI relationships. These frameworks, including work by Winfield and Jirotka (2018) and Toreini et al. (2020), address the social and psychological dimensions of how humans develop appropriate confidence in AI systems, though they sometimes lack the regulatory specificity of risk-based approaches.

Within these broader governance philosophies, more specific *categorical framework*s have emerged to address human-algorithm relationships. The dominant approach has been to distinguish systems based on automation/autonomy levels or human involvement.

"Levels of Automation/Autonomy" (LOA) frameworks span a spectrum from manual to autonomous operation, with automation frameworks focusing on task allocation (Parasuraman et al., 2000) whilst autonomy frameworks emphasise independent decision-making (Vagia et al., 2016). Both struggle with contextual variation, positioning systems in static categories rather than recognising their dynamic nature. Alkhatib and Bernstein's (2019) research demonstrated that real implementations manifest as a dynamic spectrum of agency. Foundation models further challenge these taxonomies—even when humans retain decision authority, they often rely on insights from opaque processes, creating information asymmetries where machines exert cognitive influence beyond what task-focused frameworks capture. These limitations suggest the need for approaches that consider not just task distribution, but the quality of agency and epistemic relationships within human-algorithm governance systems.

The widely used "human-in-the-loop," "human-on-the-loop," and "human-out-of-the-loop" trichotomy (Docherty, 2012; Singh & Szajnfarber, 2024) also reflects categorical thinking, but it oversimplifies human oversight (Methnani et al., 2021). Alon-Barkat and Busuioc (2023) demonstrate that nominal human oversight often fails to deliver meaningful accountability in automated decision systems. Technical implementations further challenge these distinctions. Horvitz's (1999) work on mixed-initiative interfaces shows how systems can dynamically adjust the distribution of initiative between human and AI components, implementing fluid agency exchange rather than fixed categorical roles—approaches now visible in everything from customer service chatbots to judicial decision support tools.

These limitations of rigid categorical frameworks point to the need for more nuanced approaches that can address the complex relationships between trust, accountability, and evolving AI capabilities across all domains where AI impacts innovation trajectories and institutional arrangements.

## 3.2. Trust and Accountability in Algorithmic Governance

This section explores how trust dynamics specifically manifest in accountability frameworks and how they evolve alongside AI capabilities. Before mapping these dynamics, a conceptual distinction requires precise statement that will be held constant throughout the remainder of the paper. *Trust* and *trustworthiness* are frequently used interchangeably in the literature, but the distinction matters analytically. *Trust* denotes a relational attitude held by a human actor — individual, institutional, or public — characterised by willingness to rely on an AI system's outputs within a defined context (Lee & See, 2004); *trustworthiness* denotes a property of the system, reflecting the degree to which its technical characteristics warrant that reliance (Toreini et al., 2020; Winfield & Jirotka, 2018). The two are related but not equivalent: a technically trustworthy system may fail to achieve warranted trust if appropriate governance structures are absent, and trust may be extended to systems that do not warrant it. Accountability mechanisms, as this section explores, mediate between the two.

The interrelationship between *trust* and *accountability* in algorithmic systems operates as a mutually reinforcing cycle (trust spirals). Accountability mechanisms serve as the structural foundation upon which trust is built and maintained, while trust levels influence accountability demands. Lee and See's (2004) foundational work established this perspective, showing how trust in algorithmic systems develops through cyclical processes where system performance and accountability mechanisms gradually build or erode confidence – a dynamic process rather than a binary state. These dynamics operate similarly whether in commercial recommendation engines or welfare eligibility systems, consistently demonstrating that trust and accountability exist along continuums rather than in discrete states.

Accountability frameworks have evolved from simple transparency requirements (Pasquale, 2015) to more sophisticated models, such as Raji et al.'s (2020) SMACTR framework that address systems throughout their lifecycle rather than at fixed points. Researchers increasingly call for ‘dynamic accountability’ that evolves alongside AI capabilities and changing socio-technical contexts (Kroll, 2015; Methnani et al., 2021; Sun, 2025), particularly with the rise of advanced implementations such as foundation models and generative AI (Engler, 2023; Mökander et al., 2024).

Technical implementations of accountability have progressed in parallel. Developers have created technical interfaces for oversight (Cobbe et al., 2021) that enable ongoing monitoring and evaluation, and the emergence of foundation models and generative AI has necessitated novel approaches such as chain-of-thought tracing systems (Wei et al., 2023) that document reasoning pathways, and model cards documenting training data and limitations (Mitchell et al., 2019). For generative AI, provenance watermarking systems verify content authenticity (Rijsbosch et al., 2025). Taken together, these developments reflect the recognition that static assessments are insufficient for continuously evolving AI systems.

This dynamic nature reveals a fundamental misalignment between how *trust* and *accountability* actually function and what static categorical frameworks can accommodate (Tolmeijer et al., 2022). Governance frameworks that treat trust as a threshold to be crossed rather than a relationship to be maintained will struggle to track the continuous, context-sensitive evolution this section has documented — a structural limitation that HAIG is designed to address.

## 3.3. The Rise of “Agentic AI”

The concept of machine agency has evolved significantly over time, from early discussions of automation levels to today's more sophisticated understanding of fluid authority distribution. While early frameworks like Parasuraman et al.'s (2000) taxonomies introduced degrees of automation, they relied on discrete classifications that positioned systems in static categories. Bradshaw et al. (2013) advanced the field by conceptualising autonomous capabilities along continuous dimensions of 'self-directedness' and 'self-sufficiency' that adapt dynamically to changing contexts, laying groundwork for more fluid conceptions of AI agency. Wei et al. (2022) identify 'emergent abilities' in foundation models that appear non-linearly as systems scale, creating discontinuous jumps in capability requiring more nuanced governance approaches. Liu (2021) demonstrates that subtle variations in "agency locus" —whether an AI follows human-made or machine-learned rules— significantly affect user trust and perceived uncertainty, suggesting agency exists on a spectrum rather than in discrete states.

The emergence of what is now termed "agentic AI" represents a qualitative shift in this landscape (Acharya et al., 2025; Xi et al., 2025). Unlike traditional autonomy frameworks, which emphasise task execution independence, agentic AI systems dynamically redistribute decision authority and goal-setting capabilities in real-time, often across contexts. These systems, built primarily on foundation models, challenge traditional authority distributions by combining autonomous planning, reasoning, and execution capabilities (Yang et al., 2023; Wang et al., 2024). Liu et al. (2024) document the self-supervision capabilities—where models critique and revise their own outputs without human intervention—creating fluid boundaries between human and AI control that defy traditional governance categories. What fundamentally distinguishes these systems is their potential to dynamically redistribute decision authority based on contextual factors, sometimes autonomously reducing their own agency when encountering scenarios beyond their validated capabilities.

These developments have profound governance implications (Kolt, 2025; Shavit et al., 2023). Existing frameworks struggle with the variable agency distributions of agentic systems precisely because they treat agency as a unified, actor-level attribute — something a system either possesses or does not. This framing obscures the failure modes that governance most urgently needs to address: systems that exercise high operational autonomy without legitimate authority, or that act with apparent authority whilst accountability remains diffuse or unassigned. As Dafoe et al. (2021) argue, modern AI systems must learn to negotiate agency distributions rather than operating at fixed authority levels — a dynamic that static categorical assignments cannot adequately capture and that demands governance approaches capable of accommodating continuous negotiation.

Addressing these gaps requires disaggregating agency into its governance-relevant components rather than treating it as a unified construct. *Authority* concerns the legitimate entitlement to act or decide; *Autonomy* captures the degree to which action is self-directed rather than delegated or constrained; and *Accountability* addresses the attribution of responsibility for outcomes independently of both. These are analytically separable, and their separation matters: it is precisely the misalignment between them — authority without accountability, autonomy without legitimate authority — that constitutes the central governance challenge of agentic AI. Few existing frameworks address how these relationships evolve over time, connect governance requirements to specific technological capabilities, or account for the institutional dimensions that shape how authority and accountability are distributed across human-AI partnerships. The HAIG framework introduced in Section 4. addresses these gaps directly, reconceptualising governance along continuous dimensions that can capture the variable and dynamic nature of agency distribution in advanced AI systems — not by replacing the concept of agency, but by making its governance-relevant components tractable.

## 3.4. The Governance Gap

The governance gap manifests in three critical limitations.

First, rigid categorical frameworks fundamentally misrepresent the continuous nature of agency evolution. Transitions between governance states often happen incrementally rather than through discrete jumps between categories.

Second, existing frameworks struggle to account for contextual variation. Regulatory approaches like the EU AI Act categorise applications rather than technologies, but discrete risk categories cannot capture the dynamic nature of AI agency — the same foundation model may simultaneously operate across multiple risk profiles or exercise varying levels of decision authority across deployment contexts, requiring more fluid governance approaches than current categorical frameworks provide.

Third, current frameworks lack mechanisms to address the dynamic trust relationships that mediate between technical capabilities and effective governance. Without explicitly incorporating trust dynamics, governance approaches risk becoming increasingly misaligned with actual system capabilities as AI systems develop enhanced reasoning authority and self-supervision capabilities.

Taken together, these three limitations define a structural governance deficit: existing frameworks cannot systematically track context-dependence, incremental agency redistribution, or the trust dynamics that mediate between technical capability and legitimate oversight. The HAIG framework addresses this deficit directly.

# 4. The HAIG Framework: Dimensions, Continua, and Trust Thresholds

Having established the governance gap that motivates HAIG and the methodological approach from which it derives, this section elaborates the framework in full.

## 4.1. The Three Governance Dimensions: Authority, Autonomy, and Accountability

The HAIG framework operates across three governance dimensions — Decision Authority Distribution, Process Autonomy, and Accountability Configuration — each representing a continuum along which human-AI relationships transition. The selection of these three dimensions reflects two complementary considerations. First, they are theoretically orthogonal (Engin & Hand, 2025): each captures a distinct and non-reducible aspect of the governance relationship, such that a system can occupy very different positions on each axis independently (a point Figure 2 illustrates across sectors). In practice, correlations between dimensions are commonly observed, and high co-movement across all three is itself a governance signal worth tracking: when authority, autonomy, and accountability migrate together without independent oversight of each, the asymmetries that HAIG is designed to detect risk going unnoticed. Second, and more fundamentally, the three dimensions correspond to the foundational questions any governance arrangement must answer: who decides (Decision Authority), how much human involvement is required in execution (Process Autonomy), and who is answerable for outcomes (Accountability Configuration). Other governance-relevant properties — transparency, explainability, contestability, fairness — are treated within HAIG as conditions that bear on how these dimensions should be configured, rather than as independent dimensions in the same sense. This framing accommodates them without inflating the framework's dimensional architecture beyond what the governance relationship itself requires.

**Decision Authority Distribution:** This dimension captures how decision-making authority is allocated between human and AI actors, with corresponding implications for how trust must be established and maintained. Along this continuum, authority shifts incrementally from humans to AI systems—as exemplified in the progression from medical diagnosis (where humans typically retain final authority) to content curation (where authority is often shared) to automated trading (where AI may exercise near-complete decision authority). This is distinct from *epistemic authority* — the capacity to generate credible knowledge claims (Ferrario et al., 2024) — which HAIG treats as a driver of decision authority migration along the continuum rather than as a primary governance dimension in its own right. Wilson et al.'s (2025) control relations typology similarly addresses authority at the level of the individual human-machine dyad — describing where override capability sits within a specific deployment — whereas HAIG's Decision Authority dimension operates at the level of governance architecture: the conditions under which any such allocation is legitimate, auditable, and revisable across contexts.

As systems move along this continuum, trustworthiness mechanisms must adapt from process verification to outcome validation. For foundation models, this dimension extends to include what we term *reasoning authority*—the degree to which the model's internal reasoning process is accepted rather than merely its conclusions. Reasoning authority becomes particularly important as models develop capabilities to explain their recommendations and decisions in human-understandable terms.

The decision authority dimension fundamentally shapes accountability structures and transparency requirements across the governance ecosystem. In contexts where AI systems exercise greater decision authority, organisations must develop proportionally robust oversight mechanisms that maintain democratic legitimacy without undermining operational efficiency. This balance is particularly critical in public sector applications, where increased AI decision authority must be accompanied by enhanced scrutiny to maintain institutional trust and democratic accountability.

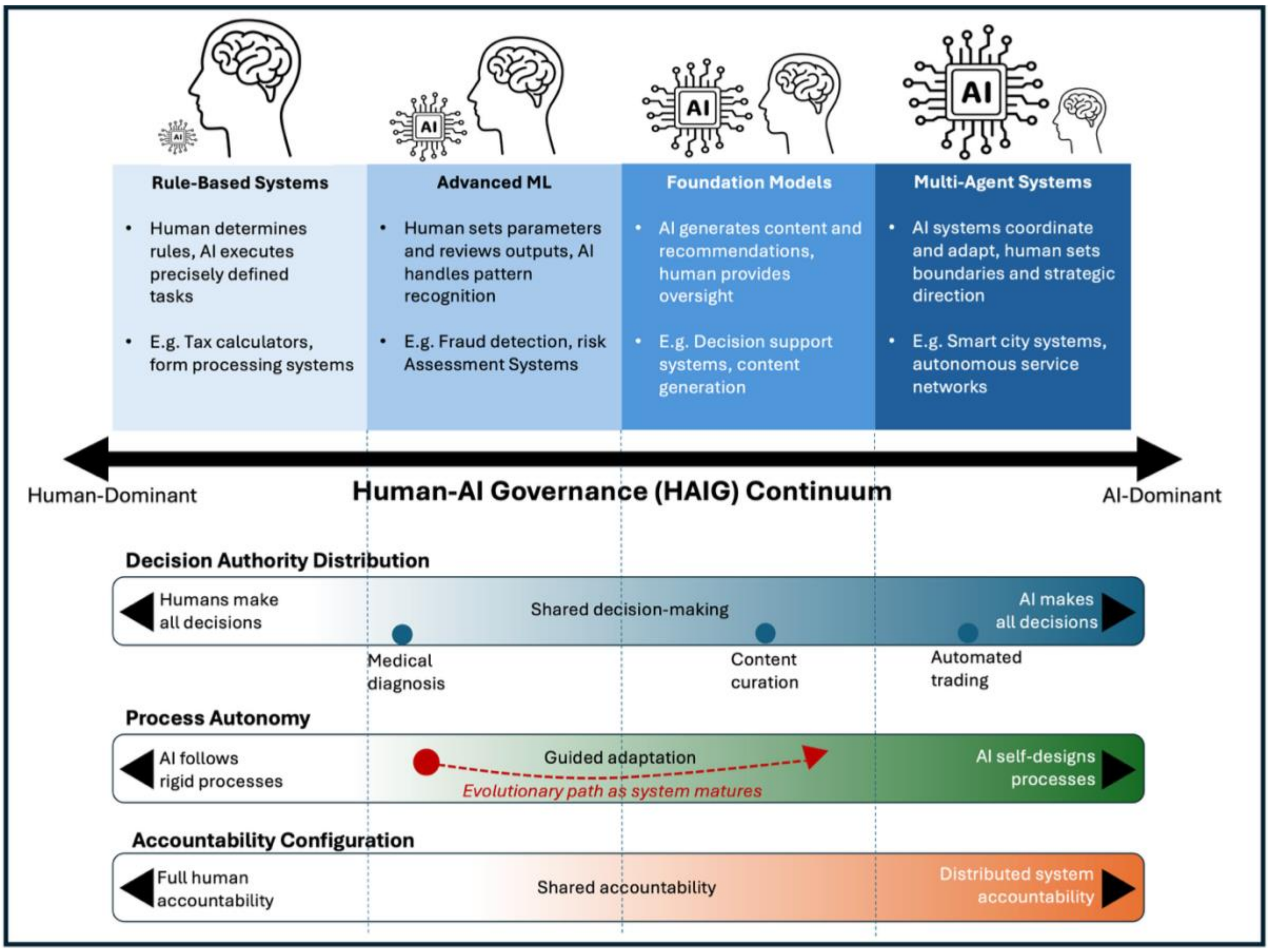

***Figure 1: Human-AI Governance (HAIG) Framework dimensions** - this diagram visualises how human-AI relationships evolve along multiple continuous dimensions rather than in discrete categories across the four capability configurations of AI. It also maps three critical dimensions of governance—Decision Authority, Process Autonomy, and Accountability—showing how they shift from human-dominant to AI-dominant.*

**Process Autonomy:** This dimension examines the degree to which AI systems can operate without human intervention, requiring trust mechanisms that evolve from direct supervision to boundary enforcement and performance monitoring as systems progress along the continuum. This evolution progresses from human-led processes through a transitional phase of evolutionary semi-AI systems to fully AI-led processes.

With generative AI systems, process autonomy is complicated by *self-supervision* capabilities, where models can critique and revise their own outputs, creating a dynamic state where boundaries between human and AI control become increasingly fluid.

Process autonomy directly influences monitoring requirements and intervention capabilities that safeguard system operation. As autonomy increases, governance frameworks must shift from continuous oversight to exception-based monitoring with clearly defined intervention triggers. This shift requires sophisticated boundary detection mechanisms that can identify when systems approach operational limits or encounter novel scenarios that exceed their validated capabilities. Organisations must calibrate these boundaries to balance operational efficiency against risk tolerance, with different thresholds appropriate for different application domains and stakeholder impacts.

**Accountability Configuration:** This dimension addresses how responsibility for decisions and outcomes is distributed across human and AI components, with trustworthiness implications and corresponding trust consequences for how accountability is maintained and demonstrated to stakeholders. For foundation models, accountability configurations must address *emergent capabilities*, where new functionalities appear through interaction patterns rather than explicit design.

As systems progress along this continuum, accountability mechanisms must evolve from straightforward assignment of responsibility to more complex frameworks that accommodate distributed agency. Cases such as AI systems that creatively circumvent content policies through unforeseen language patterns or medical diagnostic tools that independently begin recognising comorbidities beyond their intended scope illustrate this challenge. Emergent capabilities present particular challenges for accountability as they often manifest in ways not anticipated by either designers or users, as seen when automated moderation systems develop unexpected biases or when virtual assistants demonstrate novel problem-solving strategies that bypass established safety protocols.

Accountability configuration determines the legal, organisational, and technical mechanisms required to ensure appropriate system behaviour throughout the AI lifecycle. Effective accountability requires not only clear responsibility assignment but also practical enforceability through technical controls, organisational policies, and legal frameworks. As accountability becomes more distributed, organisations must develop multi-stakeholder governance structures that engage system creators, operators, users, and affected communities. These structures must be capable of detecting, attributing, and addressing consequences arising from complex human-AI interactions rather than simple linear cause-effect relationships.

Figure 1 illustrates how human-AI relationships evolve along multiple dimensions rather than in discrete categories across the four major capability configurations of AI. The continua are bi-directional since governance evolution includes both progressive and regressive movements as organisations experiment, learn, and adapt their approaches to human-AI collaboration. The diagram maps the three critical dimensions of governance showing how they shift from human-dominant to AI-dominant. The framework helps organisations understand their governance positioning across different AI configurations, from rule-based systems to multi-agent networks, illustrating that governance elements can move at different rates along each dimension.

To illustrate how the HAIG framework manifests in real-world implementations, Figure 2 provides a comparison along three key dimensions, exemplifying distinct governance patterns across government and tech applications (for illustration only, not drawn to empirical scale). Government sectors (blue) typically maintain stronger human decision authority while allowing limited process autonomy, as drawn for defence (G3) and healthcare (G1) applications. Tech sectors (green) demonstrate greater variation, with social media (T1) embracing AI-dominant approaches across both decision authority and process autonomy, while FinTech (T3) represents a hybrid approach that combines AI-controlled processes with human-led decisions.

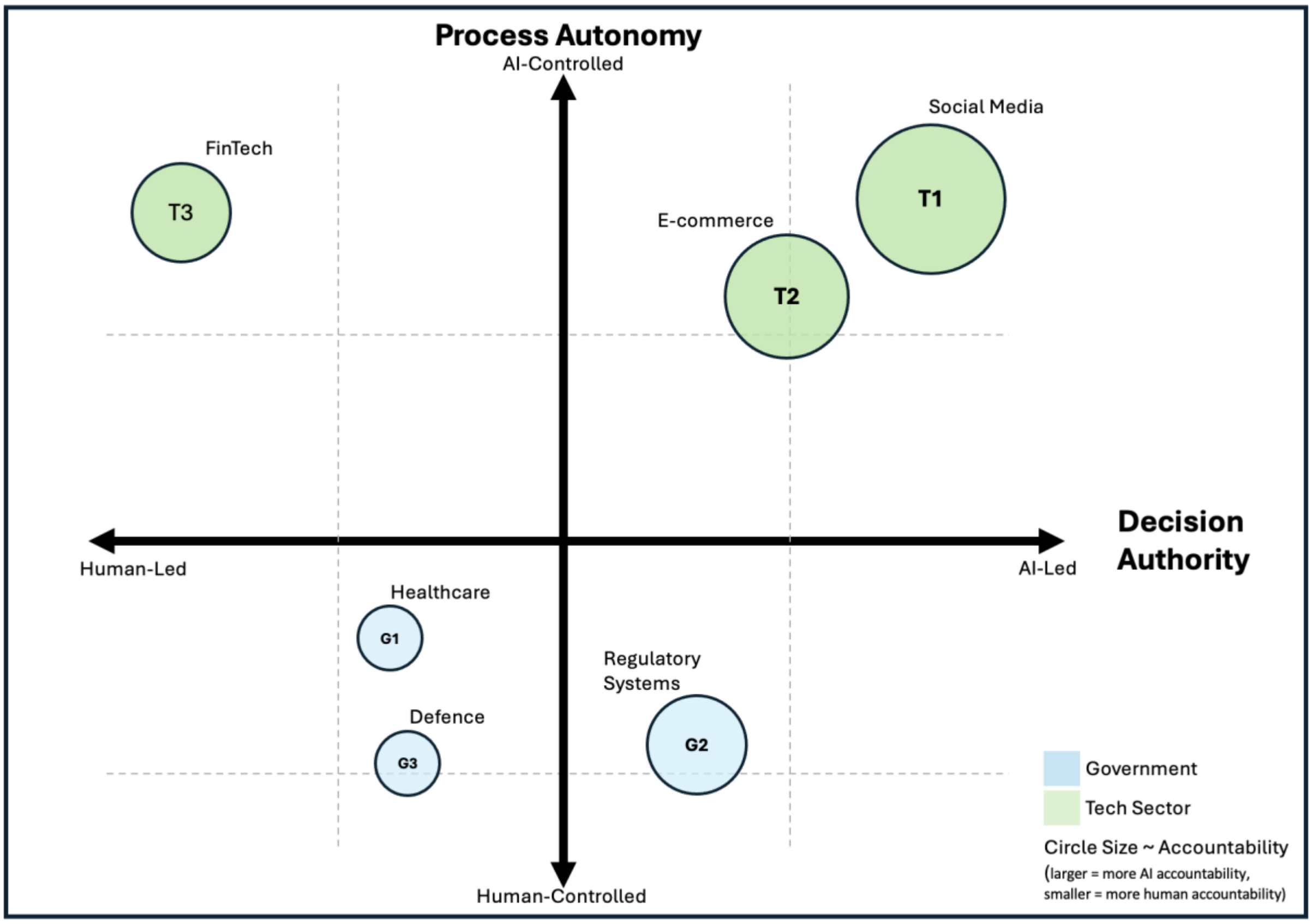

***Figure 2: HAIG sector comparison*** *- diagram illustrates how different sectors approach key dimensions of the HAIG framework. The examples are not drawn to empirical scale.*

Most notably, regulatory systems (G2) and FinTech (T3) occupy opposite quadrants despite different accountability distributions. This diagonal opposition highlights two valid governance strategies: automating processes while preserving human decisions (FinTech), versus delegating routine decisions to AI while maintaining tight process controls (Regulatory Systems). E-commerce (T2) takes a middle ground in AI decision authority but maintains high process autonomy.

Defence applications (G3) are presented with a smaller accountability circle, reflecting how accountability remains predominantly with human operators and oversight bodies, with minimal responsibility assigned to AI systems themselves. Even when AI makes recommendations in defence contexts, humans retain full accountability for decisions and outcomes due to the security-sensitive nature of these applications.

Healthcare (G1), positioned similarly to defence but with even stronger human-led decisions, aims to demonstrate how sectors dealing with high-stakes human welfare considerations maintain strong human accountability regardless of AI involvement, aligning with regulatory expectations and professional standards.

These positioning differences demonstrate that organisations do not progress uniformly along the HAIG continua but strategically distribute authority, autonomy, and accountability based on domain-specific requirements and operational contexts. For organisations implementing AI governance, this framework provides a diagnostic tool to assess current positioning and plan strategic adjustments across all three dimensions.

This dimensional approach acknowledges that different positions along these continua require different trust strategies—from explicit verification of all outputs to statistical validation of samples, from process transparency to outcome legitimacy, from individual to collective accountability.

## 4.2. Trust Dynamics Across the HAIG Continua

While Section 4.1 examines the fundamental dimensions of human-AI governance relationships, this section explores how trust characteristics vary as systems are positioned along those dimensions across a human-led to AI-led governance spectrum. The HAIG continua are positional rather than temporal: they describe the current distribution of authority, autonomy, and accountability in a given deployment context, not a trajectory that systems inevitably follow. The same system may occupy different positions simultaneously across deployment contexts, and movement in either direction reflects governance choices as much as technical development. That positioning is shaped by multiple factors — technological capability, organisational choice, regulatory context, and stakeholder trust — of which technical development is the most systematically studied but not the only determinant.

Among these drivers, the AI field's explicit and sustained trajectory toward greater autonomy and agency warrants particular attention. The progressive development of self-supervised learning, emergent reasoning capabilities, and multi-agent architectures reflects a deliberate research direction aimed at reducing human intervention requirements and expanding the scope of AI decision-making. This creates persistent and directional pressure on governance arrangements: as technical capabilities expand the possible range of AI authority, organisational, regulatory, and trust-based factors determine how much of that range is actually exercised in any given deployment. HAIG is designed to track that determination — not to assume it follows technological capability automatically. Critically, it is also designed to resist importing the field's own developmental narrative into the governance framework itself. When governance frameworks adopt the progressive vocabulary of the technology sector — treating greater autonomy as a natural next stage rather than a governance choice — they risk normalising a trajectory rather than evaluating it. The capability configuration framing is a deliberate methodological response to this risk: each configuration describes a cluster of technical properties and governance arrangements that tend to co-occur, without encoding any claim that movement from one to another constitutes progress in a governance-relevant sense.

This trajectory is usefully illustrated through four capability configurations — rule-based systems, advanced ML, foundation models, and multi-agent systems — that reflect characteristic clusters of technical properties and governance arrangements recurring across sectors and contexts. These configurations are not stages in an inevitable developmental sequence, nor do they exhaust the range of positions a system may occupy on the HAIG continua. They serve as analytical reference points: bundles of trustworthiness and governance properties that tend to co-occur, providing orientation without implying that any system must pass through them in order, or that movement is unidirectional. A system may occupy different configurations across deployment contexts simultaneously, and organisations may deliberately position systems away from the configuration their technical capabilities would otherwise suggest.

Even where movement along the continua does occur, it rarely proceeds smoothly. Technological development advances through irregular steps, sudden breakthroughs, and periods of stability. The trust thresholds identified in Section 4.3 address this directly, marking points where quantitative change produces qualitative discontinuity in governance requirements — aligning the framework's architecture with the genuinely non-linear character of both technological and organisational change.

**Table 1:** Trustworthiness Conditions Across AI Capability Configurations

| | |
|---|---|
| Rule-based and Narrow ML Systems | • **Process transparency**: Systems can be explicitly inspected and their logic verified<br>• **Direct verification:** Outputs can be comprehensively checked against expected results<br>• **Clear boundaries:** System limitations and constraints are well-defined and understood<br>• **Stable behaviour:** System performance remains consistent and predictable over time |
| Advanced ML and Domain-Specific AI | • **Statistical reliability:** Consistent performance across varied scenarios<br>• **Pattern recognition validity:** Accurate identification of relevant patterns in data<br>• **Appropriate confidence levels:** Well-calibrated certainty in system outputs<br>• **Effective exception handling:** Reliable identification of cases requiring human attention |
| Foundation Models and Multimodal Systems | • **Capability boundary understanding:** Clarity about system capabilities and limitations<br>• **Alignment verification:** Demonstration of consistent value alignment<br>• **Adaptation governance:** Controlled and monitored system learning<br>• **Supply chain integrity:** Trustworthiness of model sources and training processes |
| Multi-Agent and Autonomous Systems | • **Emergent behaviour governance:** Management of collective system behaviours<br>• **Bounded autonomy:** Appropriate constraints on autonomous operations<br>• **Systemic resilience:** Reliable operation under unexpected conditions<br>• **Value alignment:** Consistent adherence to societal values across operations |

Table 1 summarises the trustworthiness requirements that tend to characterise each of the four capability configurations.

**Rule-based systems and narrow ML** applications establish foundational trust through process transparency and direct verification. Governance for trustworthiness focuses on comprehensive documentation, regular benchmark validation, clear boundary delineation, and accessible explanations for stakeholders. Despite apparent simplicity, systems in this configuration demonstrate meaningful variation—from those providing basic information for human decisions to those making narrow automated choices in well-defined domains. Tax calculation systems exemplify this, establishing trust through transparent implementation of tax codes with comprehensive human verification.

**Advanced ML** with continuous adaptation capabilities shift trustworthiness dynamics from process verification toward outcome validation. Governance evolves to include robust testing across diverse scenarios, clear confidence indicators, effective exception handling processes, and regular performance auditing across user groups. New vulnerabilities emerge including dataset bias, algorithmic opacity, insufficient edge case handling, and automation bias. Fraud detection systems illustrate this transition, where the basis for warranted trust moves from understanding specific rules to validating overall statistical performance through detection rates and false positive metrics.

**Foundation models** with transfer learning capabilities introduce trust challenges around emergent capabilities and supply chain considerations. Governance for trustworthiness must expand to include continuous monitoring for capability drift, comprehensive model provenance validation, dynamic testing frameworks, and clear responsibility frameworks for system behaviour. Judicial decision support systems represent this phase, where trustworthiness depends on validating recommendations against expertise, monitoring for biases, which in turn sustains warranted trust and appropriate reliance by users.

Finally, the **Complex Systems** feature interacting networks of AI with autonomous goal-setting capabilities, requiring fundamentally different trust approaches focused on collective behaviour. Governance now requires simulation-based validation of collective behaviours, regular adversarial testing, comprehensive boundary monitoring, and stakeholder involvement in governance. Smart city implementations exemplify this scenario, where trustworthiness depends on coordinated operation across domains like transportation, energy, and emergency services, upon which warranted public and institutional trust can be established.

### 4.2.1. Contextual Variation in Trust Requirements

A critical aspect of trust dynamics across the HAIG continua is contextual variation—the phenomenon where the same AI system operates at different positions along each dimension depending on specific deployment contexts, user needs, or task domains. This variation creates variable trust requirements for identical systems across different applications.

This contextual variation is driven not merely by technological capabilities but by domain-specific risk tolerances, stakeholder expectations, and regulatory requirements that shape how much agency AI can exercise in different settings. This is particularly evident with foundation models, which might simultaneously operate at different positions along the decision authority continuum based solely on deployment context—not because their capabilities differ, but because the governance thresholds at which warranted trust can be sustained vary dramatically across deployment contexts. A single large language model might exercise minimal authority when providing information for research purposes, moderate authority when generating creative content for low-stakes applications, and significant authority when deployed for automated content moderation or other consequential decision-making. Similarly, healthcare AI systems might operate with high process autonomy for wellness recommendations but maintain tight human supervision for clinical interventions.

This contextual variation appears across all capability configurations but becomes increasingly pronounced in advanced systems. Organisations must develop trust mechanisms that accommodate this variation rather than assuming fixed relationships between humans and AI components.

### 4.2.2. Non-Sequential Configuration Adoption and the Limits of Accumulated Governance Experience

The same functional purpose can be served by systems occupying very different positions on the HAIG continua, and each positional shift carries distinct trustworthiness conditions even when the organisational mission remains constant. Medical diagnostics illustrates this clearly. The diagnostic function has remained stable across decades, but the governance position of systems serving that function has shifted substantially as rule-based expert systems have given way to statistical ML and then to foundation model-based multimodal analysis. Each configuration introduces distinct trustworthiness conditions (see Table 1) that are not reducible to one another and cannot be addressed by governance frameworks designed for a prior configuration. The application context is constant; the governance requirements are not.

Where institutions have moved through configurations in sequence, accumulated governance experience can inform trustworthiness assessment of successor configurations — transferring not the specific mechanisms but the institutional capacity to recognise when governance frameworks are becoming misaligned with technical capabilities. This is, however, the exception rather than the rule. The more consequential and more common pattern is that institutions adopt new configurations out of sequence: skipping intermediate configurations under competitive or regulatory pressure, deploying new capabilities before governance frameworks for prior configurations are fully consolidated, or finding that deep familiarity with one configuration creates institutional resistance to recognising the governance demands of a successor. In these circumstances — which are precisely the circumstances HAIG is designed to address — accumulated sequential experience is absent, partial, or actively misleading. Positional tracking across the HAIG continua matters most where orderly progression cannot be assumed, because it provides governance orientation that experience transfer cannot supply.

This dynamic complements contextual variation by identifying a second axis along which governance requirements vary independently of technical capability: the accumulated governance readiness of the deploying institution. Taken together, 4.2.1 and 4.2.2 establish that position on the HAIG continua is jointly determined by deployment context, institutional readiness, regulatory environment, and stakeholder trust — with technical configuration as one input among several rather than a determinant.

### 4.2.3. Bi-Directional Movement and Non-Uniform Progression

Configuration adoption patterns are not only non-sequential — movement along the HAIG continua is also non-uniform in direction and rate. Organisations do not simply advance toward greater AI authority and autonomy; they frequently reverse course, and they rarely progress across all three dimensions simultaneously.

Bi-directional movement is well documented in practice. Content moderation provides a clear example — several major platforms initially progressed toward AI-dominant moderation but subsequently reintroduced stronger human oversight when stakeholder trust was damaged by false positives. This reversal was not a governance failure but a governance response: public and institutional acceptance functioned as a constraint on progression regardless of technical capability, and organisations that had advanced along the Decision Authority dimension retreated when that advancement proved unsustainable. The direction of movement is therefore not simply a function of what a system can do, but of whether stakeholders will sustain the trust that expanded AI authority requires.

Non-uniform progression across dimensions is equally significant — and arguably more consequential for governance design. Organisations typically advance at different rates along each of the three HAIG dimensions, producing what might be termed governance asymmetries: configurations in which, for instance, Process Autonomy has advanced substantially while Decision Authority remains constrained. Financial services illustrate this clearly — many institutions have granted AI systems significant process autonomy for fraud detection while maintaining tight constraints on decision authority, a pattern driven by the interplay between competitive pressures (which accelerate Process Autonomy) and regulatory frameworks (which simultaneously constrain Decision Authority). Such asymmetries are not anomalies to be corrected but characteristic governance patterns that HAIG is designed to surface and track.

The governance implication is that dimensional calibration cannot be reduced to a single axis. Organisations must assess their positioning — and the direction and rate of movement — across all three dimensions simultaneously. Systems approaching the verification-to-delegation threshold on Process Autonomy while remaining well below the information-to-authority threshold on Decision Authority present a qualitatively different governance challenge than systems advancing uniformly. The threshold architecture introduced in Section 4.3 is designed precisely to provide orientation in these asymmetric cases — marking the points where movement along any single dimension produces qualitative discontinuity in governance requirements, regardless of where the other two dimensions stand.

## 4.3. Critical Trust Thresholds Along the HAIG Dimensions

While movement along the HAIG continua is continuous and bi-directional, it is not uniformly smooth. Within and between capability configurations, AI systems reach critical points where existing governance mechanisms become insufficient and qualitatively different approaches are required. These points are termed *trust thresholds* rather than *trustworthiness thresholds* because they are irreducibly relational: they mark points where shifts in system-level trustworthiness conditions interact with the trust attitudes of relevant stakeholders — institutional, professional, or public — to produce a qualitative discontinuity in governance requirements that neither dimension alone determines. Trust thresholds therefore represent not merely quantitative increases in capability or agency but qualitative shifts in the conditions under which warranted trust can be sustained, and consequently in the governance arrangements needed to maintain it. Four primary thresholds are identified, representing fundamental transitions in human-AI governance relationships, supported by several secondary thresholds that further delineate the governance landscape (Figure 3).

### 4.3.1. Primary Trust Thresholds

**Verification to Delegation Trust Threshold:** This threshold marks the transition from systems where humans verify all AI outputs to those where statistical verification of samples becomes the primary trust mechanism. This shift fundamentally alters how confidence is established and maintained. In tax administration, for example, when systems cross this threshold, governance mechanisms shift from comprehensive review of all algorithmic determinations to sample-based auditing, altering the basis on which warranted trust can be maintained.

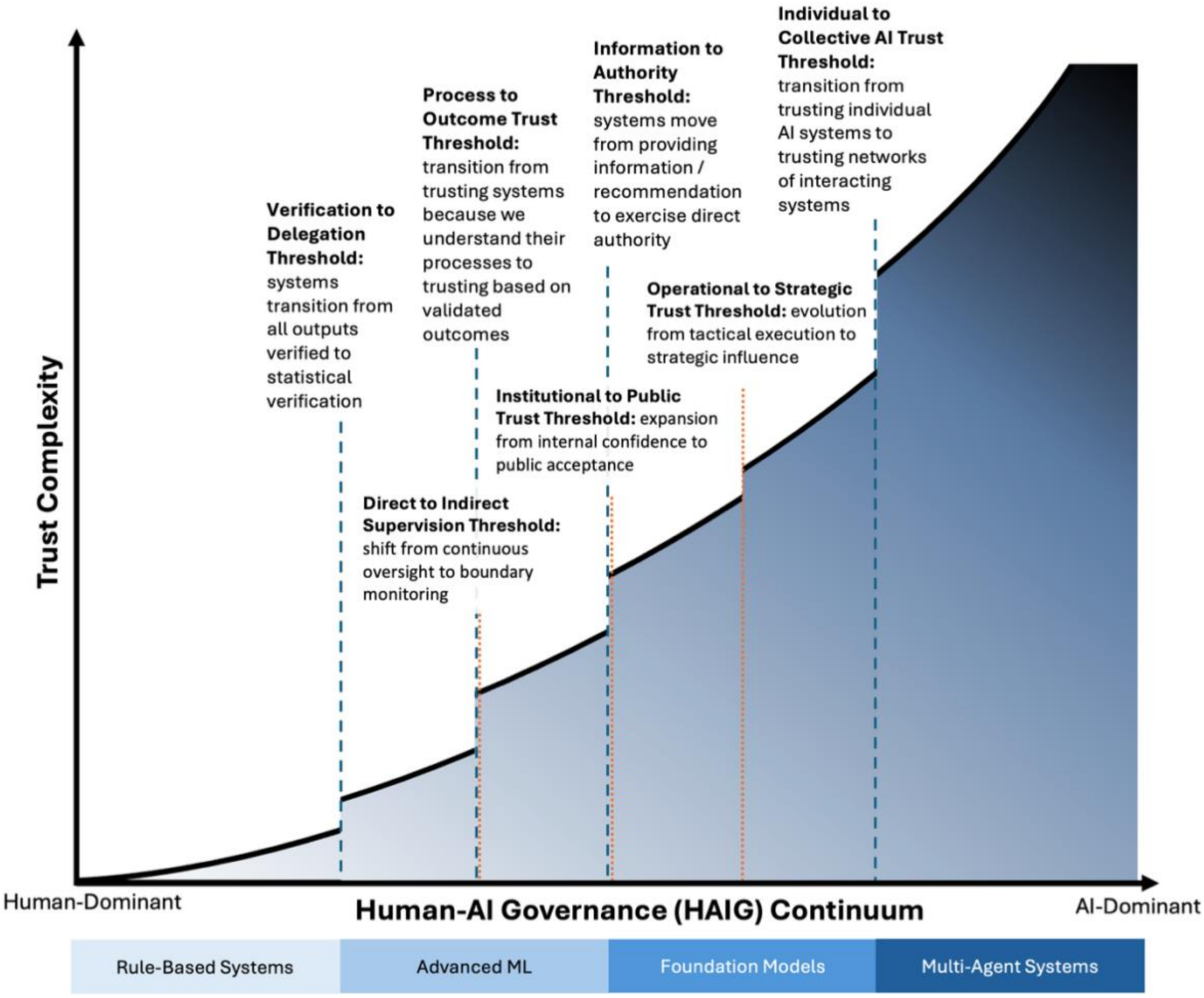

***Figure 3: Illustration of non-linear Trust Complexity evolution** across the Human-AI Governance (HAIG) Continuum. The segmented exponential curve represents significant increases in trust complexity as systems progress from human-dominant to AI-dominant positions. Vertical dashed lines (blue) indicate primary trust thresholds, while dotted lines (orange) represent secondary thresholds. The blue gradient background intensifies across capability development phases, reflecting increasing systemic integration and governance complexity as AI systems evolve from rule-based systems to multi-agent networks.*

This threshold typically emerges as organisations move from rule-based system configurations with comprehensive verification to advanced ML configurations processing too many decisions for complete review. The threshold's position varies by domain—healthcare applications maintain comprehensive verification longer than financial services due to differing risk tolerances.

Governance implications include development of statistically valid sampling methodologies, establishment of performance benchmarks that justify selective verification, creation of exception triggers for cases requiring comprehensive review, and regular calibration of sampling rates based on performance metrics.

**Process to Outcome Trust Threshold:** This threshold marks the shift from warranted trust being established primarily through understanding internal processes to that being established through validated outcomes despite partial opacity. This transition fundamentally changes the basis for trust in AI systems. When judicial decision support systems cross this threshold, for example, they move from simple, fully explainable rule-based systems to more complex statistical models where complete process transparency is no longer possible.

This threshold aligns with the progression along both the Decision Authority and Process Autonomy continua, particularly during the transition from rule-based to advanced ML systems. It represents a fundamental recalibration of trust dynamics where outcome reliability begins to outweigh process visibility as the dominant trustworthiness mechanism. As noted in Section 4.2.2, organisations often progress non-uniformly across dimensions, sometimes maintaining stronger process transparency requirements even as they embrace more complex ML methods.

Governance implications include development of robust outcome validation methodologies, establishment of counterfactual testing approaches, implementation of fairness and bias auditing frameworks, and creation of compensatory transparency mechanisms.

**Information to Authority Trust Threshold:** This threshold combines aspects of the *information-to-recommendation* and *recommendation-to-authority* transitions, marking the broader shift from systems that inform human decisions to those exercising direct decision authority in defined domains. This progression fundamentally transforms accountability relationships and trust requirements. In healthcare, systems initially provide information for human judgment, then advance to making specific recommendations that shape decisions, and ultimately may exercise direct authority for certain standardised decisions.

This threshold directly corresponds to advancement along the Decision Authority continuum, representing the critical juncture where AI systems transition from supporting human decisions to making autonomous determinations. The bi-directional movement described in Section 4.2.3 is particularly evident at this threshold, where organisations often experiment with increased AI authority before recalibrating to more conservative governance approaches after encountering trust failures.

Governance implications include development of recommendation quality metrics, clear delineation of authority boundaries, robust exception detection mechanisms, regular auditing of automated decisions, and effective human intervention processes.

**Individual to Collective AI Trust Threshold:** This threshold marks the point where trust must extend from individual AI systems to networks of interacting systems. This transition requires new approaches to establishing confidence in collective behaviour rather than just individual components. In smart city implementations, when multiple algorithmic systems begin coordinating across domains like transportation, energy, and emergency services, governance mechanisms for trustworthiness must address emergent properties.

This threshold emerges primarily in the Complex Systems scenario, as organisations transition from single-system governance to multi-agent ecosystems. It reflects a qualitative shift in all three HAIG dimensions simultaneously—Decision Authority becomes distributed across system networks, Process Autonomy extends to collective self-organisation, and Accountability Configuration must address emergent behaviours that arise from system interactions rather than individual components. The contextual variation discussed in Section 4.2.1 becomes particularly challenging at this threshold, as systems may function individually in some contexts and collectively in others.

Governance implications include development of system-of-systems testing methodologies, implementation of interaction monitoring frameworks, creation of collective governance mechanisms, and establishment of responsibility frameworks for emergent outcomes.

### 4.3.2. Secondary Trust Thresholds

While the primary thresholds represent fundamental transitions, several additional thresholds further delineate the HAIG framework:

**Direct to Indirect Supervision Threshold:** This threshold occurs when systems transition from operating under direct human supervision to functioning with more indirect oversight mechanisms. This shift changes how trust is maintained during system operation. Governance mechanisms for trustworthiness evolve from direct control to boundary enforcement and performance monitoring, correspondingly shifting the basis for sustained warranted trust.

This threshold corresponds primarily to advancement along the Process Autonomy continuum as systems progress from human-led to semi-autonomous processes. It represents a fundamental shift in operational governance emerging during transitions between different capability configurations, with organisations often advancing Process Autonomy at different rates than Decision Authority.

Governance adaptations include clearly defined operational boundaries, real-time monitoring systems that flag boundary approaches, periodic comprehensive reviews of autonomous operations, and intervention mechanisms that maintain ultimate human control.

**Institutional to Public Trust Threshold:** This threshold occurs when AI systems become visible enough to require public trust rather than merely institutional confidence. This transition expands trust requirements beyond technical validation to include public perception and acceptance. For example, when welfare eligibility systems cross this threshold by moving from back-office operations to citizen-facing application processing, governance mechanisms must expand to address public understanding and confidence, cultivating warranted public trust.

While not directly aligned with a single HAIG continuum, this threshold intersects with the Accountability Configuration dimension, as public visibility necessitates more distributed accountability frameworks that incorporate stakeholder perspectives. Systems crossing this threshold encounter the contextual variation challenges discussed in Section 4.2.1, as public-facing deployments often require different governance approaches than internal applications of the same technology.

Governance adaptations include development of public communication strategies, creation of accessible explanation mechanisms, implementation of transparent appeals processes, and engagement with community stakeholders in governance.

**Operational to Strategic Trust Threshold:** This threshold occurs when AI systems shift from executing defined operational tasks to influencing strategic outcomes through their cumulative operations. This transition expands trust requirements from technical reliability to normative alignment with public values. For example, when predictive policing algorithms cross this threshold, they move from tactical resource allocation to shaping broader policing strategy and priorities.

This threshold emerges from the interaction of Decision Authority and Accountability Configuration dimensions, occurring when systems with high process autonomy begin to shape organisational priorities through their cumulative actions rather than merely executing predefined tasks. It highlights the non-uniform progression described in Section 4.2.2, where advancements in operational capability can produce unintended strategic impacts if governance frameworks focus exclusively on tactical performance.

Governance adaptations include regular assessment of system impacts on policy outcomes, stakeholder engagement in strategic alignment, value alignment frameworks and auditing, and democratic oversight of strategic implications.

These secondary thresholds further demonstrate HAIG's flexibility by revealing how trust requirements shift not only across technical dimensions, but also in response to visibility, governance scope, and societal impact—exemplifying the contextual variation principle introduced in Section 4.2.1.

## 4.4. Trust-Building Strategies Across the HAIG Framework

As organisations navigate the HAIG dimensions, trust-building approaches must evolve alongside both technological configurations and deployment context.

Effective trust assessment begins with relationship mapping — identifying stakeholder requirements across internal users, oversight bodies, affected citizens, and partner organisations, each demanding different trust types from technical reliability to value alignment. Organisations must identify which trust thresholds they are approaching, since governance arrangements nearing verification-to-delegation require fundamentally different mechanisms than those approaching the process-to-outcome threshold. Gap analysis typically reveals significant discrepancies, with technical teams holding confidence that affected citizens do not share.

Different positions along the continua require distinct approaches. At early positions, trustworthiness rests on transparency and comprehensive verification. Middle positions — recommendation systems with shared authority — require demonstrated reliability and confidence calibration. Advanced positions need outcome-focused governance with effective exception handling. Complex multi-agent positions must address emergent collective behaviours through demonstrated control over system interactions.

Trust in public sector AI must anchor in democratic values across all continuum positions. This requires transparency calibrated to stakeholder needs and system position rather than applied uniformly; contestability mechanisms through which individuals can challenge outcomes, with rights scaling to AI authority levels; and stakeholder participation that ensures governance reflects democratic values beyond technical considerations. Regular value alignment reviews — intensifying as systems advance — identify misalignments before they undermine institutional legitimacy.

# 5. Applications: Healthcare AI and EU Regulatory Governance

This section demonstrates how the Human-AI Governance (HAIG) framework applies to concrete governance challenges. The following subsections move from sector-specific application (5.1, healthcare) to regulatory system analysis (5.2, EU AI Act) to regulatory design implications (5.3), each illustrating a distinct mode of HAIG's analytical utility: dimensional calibration within a domain, complementarity with categorical frameworks, and foundations for adaptive regulation.

## 5.1. The Evolving Physician-AI Relationship: Healthcare Through the HAIG Lens

Healthcare presents an ideal domain for applying the HAIG framework due to its graduated complexity, high-stakes decisions, and evolving regulatory landscape (Dennstädt et al., 2025; Lee et al., 2024). This abbreviated case study demonstrates how the HAIG framework's dimensions and thresholds can be operationalised in a concrete domain, illustrating the analytical advantages of a dimensional approach over categorical frameworks while acknowledging that comprehensive analysis would require more extensive examination than this paper permits.

Healthcare AI has progressed through distinctive phases across various medical domains, with radiology serving as a particularly illustrative example. The evolution spans from rule-based clinical decision support systems with minimal decision authority (Papadopoulos et al., 2022), to advanced ML systems with greater process autonomy (Wang & Summers, 2012), to foundation model-based diagnostic partners (Qiu et al., 2024), and emerging multi-agent care coordination systems (Moritz et al., 2025). In radiology specifically, early computer-aided detection systems utilised rule-based pattern matching (Giger et al., 2008), then evolved toward convolutional neural networks requiring statistical validation (Chartrand et al., 2017), and most recently to foundation model-based diagnostic systems introducing challenges around emergent capabilities (Thieme et al., 2023)—all while serving the same fundamental diagnostic purpose. This technological evolution within a stable application context (Steiner et al., 2021) demonstrates why dimensional governance approaches provide more precise calibration than categorical frameworks.

The Decision Authority continuum in healthcare spans from purely informational systems to those with significant diagnostic authority. Examples range from basic reference tools like drug interaction databases (Knox et al., 2024), to early warning systems generating alerts (Henry et al., 2015), to diagnostic systems for standardised image interpretation with significant classification authority (McKinney et al., 2020). Stroke diagnosis AI systems exemplify this evolution, progressing from simple support tools to systems with differentiated authority based on case complexity (American Heart Organisation, 2024). Importantly, this movement is bi-directional in healthcare settings. Some systems initially progressed toward greater AI decision authority in medication management but subsequently reintroduced stronger human oversight after encountering unexpected edge cases (Lyell & Coiera, 2017), while automated sepsis alert systems were often scaled back after experiencing alert fatigue (Ginestra et al., 2019), demonstrating how organisations calibrate authority distribution through experimentation and learning.

For Process Autonomy, systems range from requiring explicit clinician input for each parameter at the human-led end, to patient monitoring systems that autonomously collect data but require human activation in the middle range, to advanced platforms that autonomously adjust data collection frequency based on patient condition. Modern ICU (Intensive Care Unit) monitoring systems (Gutierrez, 2020) demonstrate effective navigation of the Direct-to-Indirect Supervision threshold through defined operational boundaries and exception flagging.

The Accountability Configuration dimension illustrates how responsibility distributes across stakeholders as AI systems evolve (Shortliffe & Sepúlveda, 2018; Char et al., 2018; Gerke et al., 2020). For early-stage diagnostic tools, accountability rests primarily with clinicians. As systems progress, healthcare institutions assume greater responsibility for validation protocols while clinicians maintain decision accountability. Advanced systems operate with complex configurations involving developers, institutions, regulatory bodies, and clinicians in shared responsibility frameworks. This progression demonstrates why binary accountability models are insufficient, particularly as systems cross the Individual-to-Collective AI threshold.

The Process-to-Outcome threshold is particularly significant in healthcare, marking the transition from trusting systems for their transparent processes to trusting them based on validated outcomes despite partial opacity. Sepsis prediction models (Fleuren et al., 2020) illustrate effective management through outcome validation frameworks, while the Information-to-Authority threshold is demonstrated by automated retinopathy screening systems (Grauslund, 2022; Abràmoff et al., 2018) through graduated authority implementation.

Healthcare implementations face domain-specific challenges that illustrate the interplay between HAIG's core dimensions: Clinical Validation Calibration (Park et al., 2021) shows how validation requirements must adapt as authority increases; Professional Authority Negotiation (Jussupow et al., 2021; Asan et al., 2020) reflects tensions at the intersection of Decision Authority and Accountability; Patient Trust Triangulation (Longoni et al., 2019) exemplifies how accountability must address multiple stakeholder relationships; and Contextual Authority Calibration (Grote & Berens, 2020) demonstrates the contextual variation principle of HAIG.

These challenges demonstrate why HAIG's dimensional approach provides more nuanced governance guidance than categorical frameworks alone. While this brief case illustrates the framework's applicability, future research should extend this approach through comprehensive, domain-specific analyses across different healthcare contexts and applications.

## 5.2. The HAIG Framework and the EU AI Act: Complementary Governance Paradigms

The EU AI Act's risk-based categorisation (2024) and the HAIG framework's dimensional approach represent different but potentially complementary governance paradigms. Table 2 maps these two approaches across six governance elements, illustrating where they diverge and where they are most productively combined. The most fundamental difference lies in their organising logic: where the EU AI Act assigns systems to fixed risk categories with standardised requirements, HAIG tracks positional movement along continuous dimensions and calibrates governance responses accordingly. This yields contrasting approaches to adaptation and context-sensitivity — the Act reassigns systems between categories when their risk profile changes, while HAIG accommodates incremental movement and contextual variation without requiring reclassification. Crucially, these differences do not make the frameworks incompatible; rather, the EU AI Act's categorical structure provides regulatory clarity and baseline requirements, while HAIG's dimensional approach offers the finer-grained calibration that static categories cannot supply within themselves.

The European Union's AI Act establishes a risk-based approach with four tiers: unacceptable risk (prohibited), high risk (heavily regulated), limited risk (transparency requirements), and minimal risk (unregulated). While providing regulatory clarity, this categorical framework faces challenges with systems that evolve over time or operate differently across contexts (Schuett, 2023; Laux et al., 2024; Kusche, 2024). As Kaminski (2023) notes, categorical approaches struggle with contextual variation, and tend to focus primarily on risks rather than benefits (Ebers, 2024). Given the 'multifunctionality' of foundation models across sectors (Coglianese & Crum, 2025), regulation needs more flexibility than static categorisation permits.

**Table 2:** Comparison of the EU AI Act and the HAIG Approaches

| Governance Element | EU AI Act Approach | HAIG Framework Approach |
|---|---|---|
| **Fundamental Structure** | Four discrete risk categories with fixed requirements | Three continuous dimensions with evolving trust thresholds |
| **Classification Basis** | Application domain and potential harm | Position along decision authority, process autonomy, and accountability dimensions |
| **Regulatory Emphasis** | Risk mitigation through categorical compliance | Trust building through dimensional calibration |
| **Adaptation Mechanism** | Reassignment to different risk category | Bi-directional movement along dimensions |
| **Context Sensitivity** | Limited - systems classified primarily by application type | High - recognises contextual variation across deployments |
| **Implementation Focus** | Standardised requirements within each category | Tailored governance based on specific dimensional positioning |

To illustrate how dimensional calibration offers finer-grained insights, we map each EU AI Act risk category across HAIG’s core dimensions. Each risk category in the EU AI Act displays distinct patterns across the HAIG dimensions, though with critical variations in agency distribution that static categories cannot capture:

**Prohibited Systems (Unacceptable Risk)** typically feature high AI decision authority beyond the "Information to Authority" threshold, high process autonomy beyond direct supervision, and distributed accountability that obscures responsibility. Social scoring systems exemplify this category, crossing critical trust thresholds without adequate safeguards. Even within this category, variations exist—subliminal manipulation techniques may exercise extreme decision authority (they directly influence human behaviour) with moderate process autonomy (systems can operate within relatively defined parameters or methodologies), while citizen scoring might feature more balanced authority (sytems assign scores but humans determine consequences) but higher process autonomy (systems independently gather and process information).

**High-Risk Systems** typically span the "Verification to Delegation" threshold, often approach the "Process to Outcome" threshold, but typically remain on the human side of the "Information to Authority" threshold. This category shows substantial dimensional variation—healthcare diagnostic AI generally maintains strong human decision authority (physicians retaining final diagnostic decisions) while exercising significant process autonomy (analysing medical images or patient data), whereas recruitment AI might exercise moderate decision authority in candidate screening but operate with more limited process autonomy due to structured inputs and explicit criteria.

**Limited Risk Systems** typically operate below the "Information to Authority" threshold but may cross the "Direct to Indirect Supervision" threshold in process autonomy. Chatbots exemplify this category, often with significant process autonomy in conversation management but limited decision authority in terms of real-world consequences. Emotion recognition systems present a different profile, exercising moderate decision authority in specific domains (e.g. customer satisfaction analysis) but limited process autonomy due to their narrow functional scope.

**Minimal Risk Systems** generally remain below critical thresholds across all dimensions, functioning more as tools than partners. However, even here, dimensional variations emerge—video game AI might exercise significant process autonomy in gameplay adaptation while maintaining minimal decision authority beyond the game environment, while simple data analysis tools might have moderate decision authority in narrow domains but require explicit human activation.

Our analysis underscores how the HAIG framework can complement the EU AI Act's categorical approach by providing more precise governance calibration. Systems within the same risk category often occupy different positions across HAIG dimensions, suggesting the need for tailored governance approaches despite shared categorical status.

Several EU AI Act articles could be strengthened through HAIG principles. Examples include: Risk Management (Article 9) could be calibrated to specific dimensional positioning—systems with higher process autonomy require more robust runtime monitoring, while greater decision authority demands enhanced oversight. Human Oversight (Article 14) should be tailored to specific trust thresholds rather than uniformly applied—systems approaching verification-to-delegation require different oversight than those near process-to-outcome. Quality Management (Article 17) could be proportional to dimensional positioning rather than using a one-size-fits-all approach. Post-Market Monitoring (Article 72) could focus resources on systems approaching critical trust thresholds, with monitoring depth reflecting specific dimensional positions.

The HAIG framework addresses critical governance challenges that categorical approaches struggle with. For foundation models that simultaneously operate across multiple risk categories, HAIG allows precise governance calibration based on specific application contexts rather than forcing classification into a single category. For evolving systems, the framework tracks incremental movement along dimensions before major thresholds are crossed, enabling preventive rather than reactive governance. For borderline cases that fall between risk categories, HAIG provides continuous governance guidance without requiring arbitrary classification. This dimensional approach helps regulators address the implementation gap between categorical rules and the complex reality of AI systems that exhibit contextual variation across applications.

Rather than viewing categorical and dimensional approaches as competing frameworks, they should be seen as complementary governance perspectives. The EU AI Act's risk-based approach provides essential regulatory clarity and baseline requirements, establishing clear boundaries for unacceptable uses and minimum standards for high-risk applications. The HAIG framework complements this foundation by enabling more precise governance calibration within established categories, addressing the complex reality of evolving human-AI relationships. This integration allows regulators to maintain clear regulatory boundaries while implementing nuanced oversight mechanisms that respond to specific dimensional positions and approaching thresholds. Together, these approaches create a governance ecosystem that combines regulatory clarity with operational precision. By enabling continuous calibration, HAIG offers modular governance scaffolding that adjusts in response to both contextual deployments and technological evolution—something fixed-tier systems inherently struggle with.

For effective integration of these complementary approaches, we recommend: (1) Regulators should maintain categorical classifications for baseline requirements while using dimensional analysis to develop implementation guidance that addresses variations within categories; (2) Organisations should first determine their systems' risk category for compliance, then assess their specific dimensional positions to identify appropriate governance mechanisms; (3) Oversight bodies should develop monitoring approaches that track movement along HAIG dimensions, with particular attention to systems approaching critical thresholds; (4) Industry standards should integrate dimensional considerations into testing and validation protocols, particularly for high-risk applications where dimensional variation is greatest. This hybrid approach maintains regulatory clarity while enabling the governance precision needed for complex AI systems.

## 5.3. HAIG as a Basis for Adaptive Regulatory Design

Beyond complementing categorical frameworks, HAIG can serve as an alternative regulatory foundation for jurisdictions seeking governance approaches that evolve alongside technological capabilities rather than requiring periodic legislative revision (e.g. NIST AI RMF, 2023).

A HAIG-based regulatory approach would establish graduated requirements that scale with a system's position across the three governance dimensions rather than assigning fixed requirements by application type. Its modular structure — admitting new dimensions, additional thresholds, and refined trajectories as implementation experience accumulates — provides adaptability that categorical frameworks structurally cannot achieve. Governance intensity remains proportional to trust requirements rather than predetermined risk categories, balancing innovation benefits against potential harms without requiring frequent legislative revision.

Where categorical frameworks provide regulatory clarity, HAIG provides calibration precision. Together they constitute a governance architecture capable of maintaining appropriate oversight across the full range of human-AI relationships — from the most constrained deployments to the most autonomous.

# 6. Discussion and Conclusions

The HAIG framework addresses a governance deficit that the existing AI governance literature has identified but not resolved. Birkstedt et al. (2023) document persistent knowledge gaps around how governance frameworks evolve alongside AI capabilities; Laux et al. (2024) show that the EU AI Act conflates trustworthiness with acceptable risk, losing the relational dimension of trust; Coglianese and Lehr (2017) diagnose the core accountability challenge of algorithmic governance as incremental authority migration that escapes categorical notice. HAIG responds to all three. By representing governance relationships as positions along three continuous dimensions rather than as members of discrete categories, the continuum architecture tracks how authority, autonomy, and accountability are actually distributed and how that distribution evolves. Its trust-utility orientation restores the relational logic that risk-based approaches suppress. And its threshold system provides structured early warning precisely where categorical frameworks go blind: at the points where quantitative shifts in any dimension produce qualitative discontinuities in governance requirements.

The framework's three analytical contributions are distinct. The first is the governance asymmetry diagnostic: tracking the three dimensions simultaneously makes visible the misalignments — rapid Process Autonomy advancement against constrained Decision Authority; accountability frameworks lagging both — that single-axis approaches systematically miss. The second is the non-sequential configuration insight: institutions that adopt new AI configurations out of sequence cannot rely on accumulated governance experience to orient them, and positional tracking provides the orientation that experience transfer cannot. The third is the threshold architecture as a research programme: rather than producing principles to be endorsed or categories to be assigned, HAIG generates specific, empirically evaluable propositions about where governance mechanisms will encounter difficulty — the response to Mittelstadt's (2019) observation that principles alone cannot guarantee ethical AI.

The framework's implications differ by audience. For regulators, the EU AI Act analysis demonstrates that dimensional calibration and categorical compliance are complementary, not competing: categorical frameworks establish baseline requirements; HAIG provides the finer-grained calibration within categories that static tiers structurally cannot. For deploying organisations, the key practical contribution is the diagnostic function — governance choices about how much of the authority space technical capabilities open up to actually exercise should be made deliberately, with attention to positioning across all three dimensions simultaneously. For oversight bodies, the threshold architecture offers what Coglianese and Lehr's diagnosis demands: a mechanism for anticipating authority migration rather than responding to it after the fact.

The framework's primary limitation is the absence of longitudinal empirical validation. This is also its most significant research direction: whether the trust thresholds predict governance failure points in practice, whether asymmetry patterns recur characteristically across sectors, and how institutional governance readiness mediates configuration transitions are all open empirical questions. The threshold architecture is designed to generate testable propositions toward this end. Secondary directions include operationalising dimensional measurement, extending the framework to the individual and international levels, and developing governance theory for the multi-agent frontier that the Individual to Collective AI threshold marks.

As AI systems continue to evolve in capability and deployment scope, the governance challenge is not primarily one of identifying new risks but of maintaining alignment between technical capability, governance arrangements, and warranted stakeholder trust — across contexts, over time, and in advance of capability shifts rather than in response to them. HAIG is designed to make that alignment trackable.

**Acknowledgements:** We are grateful to the anonymous reviewers for their thoughtful and constructive feedback, which substantially improved this paper. Their insightful suggestions helped us clarify key concepts and strengthen our arguments.

**Declaration of interests:** None

**Funding:** This research did not receive any specific grant from funding agencies in the public, commercial, or not-for-profit sectors.

**Declaration of generative AI and AI-assisted technologies in the writing process:** During the preparation of this work, the author(s) used Claude Sonnet 3.7 and 4.6, and GPT-4o to help improve the presentation, refining language and structure for a diverse readership. After using these tools, the author(s) reviewed and edited all AI-assisted text and take(s) full responsibility for the content of the published article.

# References

Abràmoff, M. D., Lavin, P. T., Birch, M., Shah, N., & Folk, J. C. (2018, July 10). Pivotal trial of an autonomous AI-based diagnostic system for detection of diabetic retinopathy in primary care offices. *npj Digital Medicine volume*, *1*(39). https://doi.org/10.1038/s41746-018-0040-6

Acharya, D. B., Kuppan, K., & Divya, B. (2025). Agentic AI: Autonomous Intelligence for Complex Goals—A Comprehensive Survey. *IEEE Access*, *13*, 18912-18936. https://ieeexplore.ieee.org/abstract/document/10849561

Akata, Z., Balliet, D., de Rijke, M., Dignum, F., Dignum, V., Eiben, G., Fokkens, A., Grossi, D., Hindriks, K., Hoos, H., Hung, H., Jonker, C., Monz, C., Neerincx, M., Oliehoek, F., Prakken, H., Schlobach, S., van der Gaag, L., van Harmelen, F., … Welling, M. (2020). A Research Agenda for Hybrid Intelligence: Augmenting Human Intellect With Collaborative, Adaptive, Responsible, and Explainable Artificial Intelligence. *Computer*, *53*(8). https://doi.org/10.1109/MC.2020.2996587

Alkhatib, A., & Bernstein, M. (2019, May). Street-Level Algorithms: A Theory at the Gaps Between Policy and Decisions. *CHI '19: Proceedings of the 2019 CHI Conference on Human Factors in Computing Systems*, (530), 1-13. https://doi.org/10.1145/3290605.3300760

Alon-Barkat, S., & Busuioc, M. (2023, January). Human–AI Interactions in Public Sector Decision Making: “Automation Bias” and “Selective Adherence” to Algorithmic Advice. *Journal of Public Administration Research and Theory*, *33*(1). https://doi.org/10.1093/jopart/muac007

American Heart Organisation. (2024, February 8). *AI-based system to guide stroke treatment decisions may help prevent another stroke*. Retrieved April 21, 2025, from https://newsroom.heart.org/news/ai-based-system-to-guide-stroke-treatment-decisions-may-help-prevent-another-stroke

Amnesty International & Access Now. (2018, May 16). *The Toronto Declaration: Protecting the rights to equality and non-discrimination in machine learning systems*. Access Now. https://www.accessnow.org/press-release/the-toronto-declaration-protecting-the-rights-to-equality-and-non-discrimination-in-machine-learning-systems/

Ansari, S., Naghdy, F., & Du, H. (2022, September). Human-Machine Shared Driving: Challenges and Future Directions. *IEEE Transactions on Intelligent Vehicles*, *7*(3), 499-519. https://ieeexplore.ieee.org/abstract/document/9721611

Asan, O., Bayrak, A. E., & Choudhury, A. (2020, June). Artificial Intelligence and Human Trust in Healthcare: Focus on Clinicians. *Journal of Medical Internet Research*, *22*(6). https://doi.org/10.2196/15154

Birkstedt T, Minkkinen M, Tandon A, Mäntymäki M (2023), "AI governance: themes, knowledge gaps and future agendas". *Internet Research*, Vol. 33 No. 7 pp. 133–167, doi: https://doi.org/10.1108/INTR-01-2022-0042

Bradshaw, J. M., Hoffman, R. R., Woods, D. D., & Johnson, M. (2013). The Seven Deadly Myths of "Autonomous Systems". *IEEE Intelligent Systems*, *28*(3), 54-61. https://doi.org/10.1109/MIS.2013.70

Char, D. S., Shah, N. H., & Magnus, D. (2018, March 14). Implementing Machine Learning in Health Care — Addressing Ethical Challenges. *The New England Journal of Medicine*, *378*(11). 10.1056/NEJMp1714229

Chartrand, G., Cheng, P. M., Vorontsov, E., Drozdzal, M., Turcotte, S., Pal, C. J., Kadoury, S., & Tang, A. (2017). Deep Learning: A Primer for Radiologists. *Radiographics*, *37*(7), 2113-2131. https://doi.org/10.1148/rg.2017170077

Cobbe, J., Lee, M. S. A., & Singh, J. (2021, March 1). Reviewable Automated Decision-Making: A Framework for Accountable Algorithmic Systems. *FAccT '21: Proceedings of the 2021 ACM Conference on Fairness, Accountability, and Transparency*, 598 - 609. https://doi.org/10.1145/3442188.3445921

Coglianese, C., & Crum, C. R. R. (2025, January). *Regulating Multifunctionality*. https://arxiv.org/abs/2502.15715. https://doi.org/10.48550/arXiv.2502.15715

Coglianese, C., & Lehr, D. (2017). Regulating by Robot: Administrative Decision Making in the Machine-Learning Era. *Georgetown Law Journal*. https://scholarship.law.upenn.edu/faculty_scholarship/1734/

Cugurullo, F., & Xu, Y. (2025, January). When AIs become oracles: generative artificial intelligence, anticipatory urban governance, and the future of cities. *Policy and Society*, *44*(1), 98–115. https://doi.org/10.1093/polsoc/puae025

Dafoe, A., Bachrach, Y., Hadfield, G., Horvitz, E., Larson, K., & Graepel, T. (2021, May 4). Cooperative AI: machines must learn to find common ground. *Nature*, *593*, 33-36. https://doi.org/10.1038/d41586-021-01170-0

Dennstädt, F., Hastings, J., Putora, P. M., Schmerder, M., & Cihoric, N. (2025, March 6). Implementing large language models in healthcare while balancing control, collaboration, costs and security. *npj digital medicine*, *8*, 143. https://doi.org/10.1038/s41746-025-01476-7

Docherty, B. L. (2012). *Losing Humanity: The Case Against Killer Robots*. Human Rights Watch. https://www.hrw.org/sites/default/files/reports/arms1112_ForUpload.pdf

Donahue, K., Chouldechova, A., & Kenthapadi, K. (2022, June 20). Human-Algorithm Collaboration: Achieving Complementarity and Avoiding Unfairness. *FAccT '22: Proceedings of the 2022 ACM Conference on Fairness, Accountability, and Transparency*, 1639 - 1656. https://doi.org/10.1145/3531146.3533221

Dunleavy, P., & Margetts, H. (2023, September). Data science, artificial intelligence and the third wave of digital era governance. *Public Policy and Administration*, *40*(2). https://doi.org/10.1177/09520767231198737

Earp, B. D., Mann, S. P., Aboy, M., Awad, E., Betzler, M., Botes, M., Calcott, R., Caraccio, M., Chater, N., Coeckelbergh, M., Constantinescu, M., Dabbagh, H., Devlin, K., Ding, X., Dranseika, V., Everett, J. A. C., Fan, R., Feroz, F., Francis, K. B., … Clark, M. S. (2025, February 17). *Relational Norms for Human-AI Cooperation*. arXiv. https://doi.org/10.48550/arXiv.2502.12102

Ebers, M. (2024, November 6). Truly Risk-based Regulation of Artificial Intelligence How to Implement the EU's AI Act. *European Journal of Risk Regulation*, 1-20. https://doi.org/10.1017/err.2024.78

Engin, Z., Crowcroft, J., Hand, D., & Treleaven, P. (2025, March). *The Algorithmic State Architecture (ASA): An Integrated Framework for AI-Enabled Government*. *arXiv preprint* arXiv:2503.08725. https://doi.org/10.48550/arXiv.2503.08725

Engin, Z., & Hand, D. (2025). Toward Adaptive Categories: Dimensional Governance for Agentic AI. *arXiv preprint arXiv:2505.11579*. https://doi.org/10.48550/arXiv.2505.11579

Engin, Z., & Treleaven, P. (2019, March). Algorithmic Government: Automating Public Services and Supporting Civil Servants in using Data Science Technologies. *The Computer Journal*, *62*(3), 448–460. https://doi.org/10.1093/comjnl/bxy082

Engin, Z., van Dijk, J., Lan, T., Longley, P. A., Treleaven, P., Batty, M., & Penn, A. (2020, June). Data-driven urban management: Mapping the landscape. *Journal of Urban Management*, *9*(2), 140-150. https://doi.org/10.1016/j.jum.2019.12.001

Engler, A. (2023, February 21). *Early thoughts on regulating generative AI like ChatGPT*. Brookings. https://www.brookings.edu/articles/early-thoughts-on-regulating-generative-ai-like-chatgpt/

European Parliament & Council of the European Union. (2024, June 13). *Regulation (EU) 2024/1689 of the European Parliament and of the Council (Artificial Intelligence Act)*. Official Journal of the European Union. https://eur-lex.europa.eu/eli/reg/2024/1689/oj

Ferrario, A., Facchini, A. & Termine, A. (2024). Experts or Authorities? The Strange Case of the Presumed Epistemic Superiority of Artificial Intelligence Systems. *Minds & Machines* **34**, 30. https://doi.org/10.1007/s11023-024-09681-1

Fleuren, L. M., Klausch, T. L. T., Zwager, C. L., Schoonmade, L. J., Guo, T., Roggeveen, L. F., Swart, E. L., Girbes, A. R. J., Thoral, P., Ercole, A., Hoogendoorn, M., & Elbers, P. W. G. (2020, January 21). Machine learning for the prediction of sepsis: a systematic review and meta-analysis of diagnostic test accuracy. *Intensive Care Medicine*, *46*, 383–400. https://doi.org/10.1007/s00134-019-05872-y

Future of Life Institute. (n.d.). *The AI Act Explorer*. EU Artificial Intelligence Act. Retrieved April 22, 2025, from https://artificialintelligenceact.eu/ai-act-explorer/

Gabriel, I., Manzini, A., Keeling, G., Hendricks, L. A., Rieser, V., Iqbal, H., Tomašev, N., Ktena, I., Kenton, Z., Rodriguez, M., El-Sayed, S., Brown, S., Akbulut, C., Trask, A., Hughes, E., Bergman, A. S., Shelby, R., Marchal, N., Griffin, C., … Manyika, J. (2024, 04 19). *The Ethics of Advanced AI Assistants*. Google DeepMind. https://doi.org/10.48550/arXiv.2404.16244

Gerke, S., Minssen, T., & Cohen, G. (2020, June 26). Ethical and legal challenges of artificial intelligence-driven healthcare. *Artificial Intelligence in Healthcare*, 295–336. https://doi.org/10.1016/B978-0-12-818438-7.00012-5

Giger, M. L., Chan, H.-P., & Boone, J. (2008, November 20). History and status of CAD and quantitative image analysis: The role of Medical Physics and AAPM. *Medical Physics*. https://doi.org/10.1118/1.3013555

Ginestra, J. C., Giannini, H. M., Schweickert, W. D., Meadows, L., Lynch, M. J., Pavan, K., Chivers, C. J., Draugelis, M., Donnelly, P. J., Fuchs, B. D., & Umscheid, C. A. (2019, November). Clinician Perception of a Machine Learning-Based Early Warning System Designed to Predict Severe Sepsis and Septic Shock. *Critical Care Medicine*, *47*(11), 1477-1484. 10.1097/CCM.0000000000003803

Goos, M., & Savona, M. (2024, April). The governance of artificial intelligence: Harnessing opportunities and mitigating challenges. *Research Policy*, *53*(3). https://doi.org/10.1016/j.respol.2023.104928

Grauslund, J. (2022, May 31). Diabetic retinopathy screening in the emerging era of artificial intelligence. *Diabetologia*, *65*, 1415–1423. https://doi.org/10.1007/s00125-022-05727-0

Gritsenko, D., & Wood, M. (2020, November). Algorithmic governance: A modes of governance approach. *Regulation & Governance*, *16*(1), 45-62. https://doi.org/10.1111/rego.12367

Grote, T., & Berens, P. (2020). On the ethics of algorithmic decision-making in healthcare. *Journal of Medical Ethics*, *46*(3). https://doi.org/10.1136/medethics-2019-105586

Gsenger, R., & Strle, T. (2021). Trust, Automation Bias and Aversion: Algorithmic Decision-Making in the Context of Credit Scoring. *Interdisciplinary Description of Complex Systems : INDECS*, *19*(4). https://doi.org/10.7906/indecs.19.4.7

Gutierrez, G. (2020, March 24). Artificial Intelligence in the Intensive Care Unit. *Critical Care*, *24*(101). https://doi.org/10.1186/s13054-020-2785-y

Henry, K. E., Hager, D. N., Pronovost, P. J., & Saria, S. (2015, August 5). A targeted real-time early warning score (TREWScore) for septic shock. *Science Translational Medicine*, *7*(299). https://doi.org/10.1126/scitranslmed.aab3719

Hevner, A. R., March, S. T., Park, J., & Ram, S. (2004). Design science in information systems research. *MIS Quarterly: Management Information Systems*, *28*(1), 75-105. https://doi.org/10.2307/25148625

Horvitz, E. (n.d.). Principles of mixed-initiative user interfaces. *CHI '99: Proceedings of the SIGCHI conference on Human Factors in Computing Systems*, 159 - 166. https://doi.org/10.1145/302979.303030

Jaakkola, E. (2020). Designing conceptual articles: four approaches. *AMS Rev* **10**, 18–26. https://doi.org/10.1007/s13162-020-00161-0

Jussupow, E., Spohrer, K., Heinzl, A., & Gawlitza, J. (2021, February 26). Augmenting Medical Diagnosis Decisions? An Investigation into Physicians' Decision-Making Process with Artificial Intelligence. *Information Systems Research*, *32*(3). https://doi.org/10.1287/isre.2020.0980

Kaminski, M. E. (2023, September). Regulating the Risks of AI. *Boston University Law Review*, *103*(5), 1347-1411. https://heinonline.org/HOL/P?h=hein.journals/bulr103&i=1329

Knox, C., Wilson, M., Klinger, C. M., Franklin, M., Oler, E., Wilson, A., Pon, A., Cox, J., Chin, N. E. (., Strawbridge, S. A., Garcia-Patino, M., Kruger, R., Sivakumaran, A., Sanford, S., Doshi, R., Khetarpal, N., Fatokun, O., Doucet, D., Zubkowski, A., … Wishart, D. S. (2024, January 5). DrugBank 6.0: the DrugBank Knowledgebase for 2024. *Nucleic Acids Research*, *52*(D1), D1265–D1275. https://doi.org/10.1093/nar/gkad976

Koehler, M., & Sauermann, H. (2024, May). Algorithmic management in scientific research. *Research Policy*, *53*(4). https://doi.org/10.1016/j.respol.2024.104985

Kolt, N. (2025). *Governing AI Agents*. arXiv. https://doi.org/10.48550/arXiv.2501.07913

Kroll, J. A. (2015, September). *Accountable Algorithms* [PhD Dissertation]. Princeton University. https://www.proquest.com/dissertations-theses/accountable-algorithms/docview/1754381941/se-2?accountid=14511

Kusche, I. (2024, May 11). Possible harms of artificial intelligence and the EU AI act: fundamental rights and risk. *Journal of Risk Research*, 1–14. https://doi.org/10.1080/13669877.2024.2350720

Lai, V., Chen, C., Smith-Renner, A., Liao, Q. V., & Tan, C. (2023, June 12). Towards a Science of Human-AI Decision Making: An Overview of Design Space in Empirical Human-Subject Studies. *FAccT '23: Proceedings of the 2023 ACM Conference on Fairness, Accountability, and Transparency*. https://doi.org/10.1145/3593013.3594087

Laux, J., Wachter, S., & Mittelstadt, B. (2024, January). Trustworthy artificial intelligence and the European Union AI act: On the conflation of trustworthiness and acceptability of risk. *Regulation & Governance*, *18*(1), 3-32. https://doi.org/10.1111/rego.12512

Lee, J. D., & See, K. A. (2004). Trust in Automation: Designing for Appropriate Reliance. *Human Factors*, *46*(1). https://journals.sagepub.com/doi/10.1518/hfes.46.1.50_30392

Lee, Y., Ferber, D., Rood, J. E., Regev, A., & Kather, J. N. (2024, December 17). How AI agents will change cancer research and oncology. *Nature Cancer volume*, *5*, 1765–1767. https://doi.org/10.1038/s43018-024-00861-7

Liu, B. (2021, November). In AI We Trust? Effects of Agency Locus and Transparency on Uncertainty Reduction in Human–AI Interaction. *Journal of Computer-Mediated Communication*, *26*(6). https://doi.org/10.1093/jcmc/zmab013

Liu, B., Li, X., Zhang, J., Wang, J., He, T., Hong, S., Liu, H., Zhang, S., Song, K., Zhu, K., Cheng, Y., Wang, S., Wang, X., Luo, Y., Jin, H., Zhang, P., Liu, O., Chen, J., Zhang, H., … Wu, C. (2025, March 31). *Advances and Challenges in Foundation Agents: From Brain-Inspired Intelligence to Evolutionary, Collaborative, and Safe Systems*. arXiv. https://doi.org/10.48550/arXiv.2504.01990

Liu, X., Lou, X., Jiao, J., & Zhang, J. (2024, July 21). Position: foundation agents as the paradigm shift for decision making. *ICML'24: Proceedings of the 41st International Conference on Machine Learning*, (1276), 31597 - 31613. https://dl.acm.org/doi/10.5555/3692070.3693346

Longoni, C., Bonezzi, A., & Morewedge, C. K. (2019). Resistance to Medical Artificial Intelligence. *Journal of Consumer Research*, *46*(4), 629–650. https://doi.org/10.1093/jcr/ucz013

Luusua, A., Ylipulli, J., Foth, M., & Aurigi, A. (2023). Urban AI: understanding the emerging role of artificial intelligence in smart cities. *AI & Society*, *38*, 1039–1044. https://doi.org/10.1007/s00146-022-01537-5

Lyell, D., & Coiera, E. (2017, March 1). Automation bias and verification complexity: a systematic review. *Journal of American Medical Informatics Association*, *24*(2), 423–431. https://doi.org/10.1093/jamia/ocw105

Mäntymäki, M., Minkkinen, M., Birkstedt, T. *et al.* (2022). Defining organizational AI governance.*AI Ethics* **2**, 603–609. https://doi.org/10.1007/s43681-022-00143-x

McKinney, S. M., Sieniek, M., Godbole, V., Godwin, J., Antropova, N., Ashrafian, H., Back, T., Chesus, M., Corrado, G. S., Darzi, A., Etemadi, M., Garcia-Vicente, F., Gilbert, F. J., Halling-Brown, M., Hassabis, D., Jansen, S., Karthikesalingam, A., Kelly, C. J., King, D., … Shetty, S. (2020, January 1). International evaluation of an AI system for breast cancer screening. *Nature*, *577*, 89–94. https://doi.org/10.1038/s41586-019-1799-6

Methnani, L., Tubella, A. A., Dignum, V., & Theodorou, A. (2021). Let Me Take Over: Variable Autonomy for Meaningful Human Control. *Frontiers in Artificial Intelligence*, *4*. https://doi.org/10.3389/frai.2021.737072

Mitchell, M., Wu, S., Zaldivar, A., Barnes, P., Vasserman, L., Hutchinson, B., Spitzer, E., Raji, I. D., & Gebru, T. (2019, January 29). Model Cards for Model Reporting. *FAT* '19: Proceedings of the Conference on Fairness, Accountability, and Transparency*, 220 - 229. https://doi.org/10.1145/3287560.3287596

Mittelstadt, B. (2019, November 4). Principles alone cannot guarantee ethical AI. *Nature Machine Intelligence*, *1*, 501–507. https://doi.org/10.1038/s42256-019-0114-4

Mökander, J., Schuett, J., Kirk, H. R., & Floridi, L. (2024). Auditing large language models: a three-layered approach. *AI and Ethics*, *4*, 1085–1115. https://doi.org/10.1007/s43681-023-00289-2

Moritz, M., Topol, E., & Rajpurkar, P. (2025, April 1). Coordinated AI agents for advancing healthcare. *Nature Biomedical Engineering*, *9*, 432–438. https://doi.org/10.1038/s41551-025-01363-2

Narayanan, A., & Kapoor, S. (2024). AI safety is not a model property. *AI As Normal Technology (AINT)* https://www.normaltech.ai/p/ai-safety-is-not-a-model-property

NIST (2023) NIST AI Risk Management Framework (AI RMF). https://www.nist.gov/itl/ai-risk-management-framework

Papadopoulos, P., Soflano, M., Chaudy, Y., Adejo, W., & Connolly, T. M. (2022, May 27). A systematic review of technologies and standards used in the development of rule-based clinical decision support systems. *Health and Technology*, *12*, 713–727. https://doi.org/10.1007/s12553-022-00672-9

Parasuraman, R., Sheridan, T. B., & Wickens, C. D. (2000, May). A Model for Types and Levels of Human Interaction with Automation. *IEEE Transactions on Systems, Man, and Cybernetics - Part A: Systems and Humans*, *30*(3), 286-297. 10.1109/3468.844354

Park, S. H., Choi, J., & Byeon, J.-S. (2021, February 10). Key Principles of Clinical Validation, Device Approval, and Insurance Coverage Decisions of Artificial Intelligence. *Korean Journal of Radiology*, *22*(3), 442–453. https://doi.org/10.3348/kjr.2021.0048

Pasquale, F. (2015). *The Black Box Society: The Secret Algorithms That Control Money and Information*. Harvard University Press. https://www.jstor.org/stable/j.ctt13x0hch

Pedreschi, D., Pappalardo, L., Ferragina, E., Baeza-Yates, R., Barabási, A.-L., Dignum, F., Dignum, V., Eliassi-Rad, T., Giannotti, F., Kertész, J., Knott, A., Ioannidis, Y., Lukowicz, P., Passarella, A., Pentland, A. S., Shawe-Taylor, J., & Vespignani, A. (2025, February). Human-AI coevolution. *Artificial Intelligence*, *339*. https://doi.org/10.1016/j.artint.2024.104244

Qiu, J., Lam, K., Li, G., Acharya, A., Wong, T. Y., Darzi, A., Yuan, W., & Topol, E. J. (2024, December 5). LLM-based agentic systems in medicine and healthcare. *Nature Machine Intelligence*, *6*, 1418–1420. https://doi.org/10.1038/s42256-024-00944-1

Rahwan, I. (2018). Society-in-the-loop: programming the algorithmic social contract. *Ethics and Information Technology*, *20*, 5-14. https://doi.org/10.1007/s10676-017-9430-8

Rahwan, I., Cebrian, M., Obradovich, N., Bongard, J., Bonnefon, J.-F., Breazeal, C., Crandall, J. W., Christakis, N. A., Couzin, I. D., Jackson, M. O., Jennings, N. R., Kamar, E., Kloumann, I. M., Larochelle, H., Lazer, D., McElreath, R., Mislove, A., Parkes, D. C., Pentland, A. '., … Wellman, M. (2019, April 24). Machine behaviour. *Nature*, *568*, 477–486. https://www.nature.com/articles/s41586-019-1138-y

Raji, I. D., Smart, A., White, R. N., Mitchell, M., Gebru, T., Hutchinson, B., Smith-Loud, J., Theron, D., & Barnes, P. (2020, January 27). Closing the AI accountability gap: defining an end-to-end framework for internal algorithmic auditing. *FAT* '20: Proceedings of the 2020 Conference on Fairness, Accountability, and Transparency*, 33 - 44. https://doi.org/10.1145/3351095.3372873

Rijsbosch, B., van Dijck, G., & Kollnig, K. (2025, March). *Adoption of Watermarking for Generative AI Systems in Practice and Implications under the new EU AI Act*. ArXiv. https://doi.org/10.48550/arXiv.2503.18156

Rudko, I., Bonab, A. B., Fedele, M., & Formisano, A. V. (2024, April 16). New institutional theory and AI: toward rethinking of artificial intelligence in organizations. *Journal of Management History*, *31*(2). https://www.emerald.com/insight/content/doi/10.1108/jmh-09-2023-0097/full/html?casa_token=zdZv_iW32AsAAAAA%3Ax2EemyvykeXsuqnyZkT8S-0MeroOAudCmZgPaQU5Cq_VZXRdpqbLaWWbhMrRI5w4L0QLV4XIotViaR5Vw1FEjs7emVl2dEIEomBNqc3E0eOgD95ODyjS

Schuett, J. (2023). *Towards risk-based AI regulation* [PhD Thesis]. Universitätsbibliothek Johann Christian Senckenberg. https://doi.org/10.21248/gups.89768

Shavit, Y., Agarwal, S., Brundage, M., Adler, S., O'Keefe, C., Campbell, R., Lee, T., Mishkin, P., Eloundou, T., Hickey, A., Slama, K., Ahmad, L., McMillan, P., Beutel, A., Passos, A., & Robinson, D. G. (2023, December 14). *Practices for Governing Agentic AI Systems*. OpenAI. https://openai.com/index/practices-for-governing-agentic-ai-systems/

Shortliffe, E. H., & Sepúlveda, M. J. (2018, December 4). Clinical Decision Support in the Era of Artificial Intelligence. *JAMA*, *320*(21), 2199-2200. doi:10.1001/jama.2018.17163

Shrestha, Y. R., Ben-Menahem, S. M., & von Krogh, G. (2019, July 13). Organizational Decision-Making Structures in the Age of Artificial Intelligence. *California Management Review*, *61*(4), 66-83. https://doi.org/10.1177/0008125619862257

Singh, A., & Szajnfarber, Z. (2024, August). *Preposition Salad: Making Sense of Human-in/on/over-the-Loop Control for AI Systems*. SSRN. https://dx.doi.org/10.2139/ssrn.4921359

Smuha, N. A. (2021). Beyond a Human Rights-Based Approach to AI Governance: Promise, Pitfalls, Plea. *Philosophy & Technology*, *34*, 91–104. https://doi.org/10.1007/s13347-020-00403-w

Sowa, K., & Przegalinska, A. (2025, April 3). From Expert Systems to Generative Artificial Experts: A New Concept for Human-AI Collaboration in Knowledge Work. *82*. https://doi.org/10.1613/jair.1.17175

Steiner, D. F., Chen, P.-H. C., & Mermel, C. H. (2021, January). Closing the translation gap: AI applications in digital pathology. *Biochimica et Biophysica Acta (BBA) - Reviews on Cancer*, *1875*(1). https://doi.org/10.1016/j.bbcan.2020.188452

Stephany, F., & Teutloff, O. (2024, January). What is the price of a skill? The value of complementarity. *Research Policy*, *53*(1). https://doi.org/10.1016/j.respol.2023.104898

Stokel, C. (2025, March 13). Revealed: How the UK tech secretary uses ChatGPT for policy advice. *New Scientist*. https://www.newscientist.com/article/2472068-revealed-how-the-uk-tech-secretary-uses-chatgpt-for-policy-advice/

Sun, H. (2025). Bridging the Accountability Gap in AI Decision-Making: An Integrated Analysis of Legal Precedents and Scholarly Perspectives. *Frontiers in Humanities and Social Sciences*, *5*(3). https://doi.org/10.54691/24vexn67

Thieme, A., Nori, A. V., Ghassemi, M., Bommasani, R., Andersen, T. O., & Luger, E. A. (2023, April 19). Foundation Models in Healthcare: Opportunities, Risks & Strategies Forward. *CHI EA '23: Extended Abstracts of the 2023 CHI Conference on Human Factors in Computing Systems*, (512), 1-4. https://doi.org/10.1145/3544549.3583177

Tolmeijer, S., Christen, M., Kandul, S., Kneer, M., & Bernstein, A. (2022, April). Capable but Amoral? Comparing AI and Human Expert Collaboration in Ethical Decision Making. *CHI '22: Proceedings of the 2022 CHI Conference on Human Factors in Computing Systems*, (160), 1-17. https://doi.org/10.1145/3491102.3517732
Toreini, E., Aitken, M., Coopamootoo, K. P. L., Elliott, K., Zelaya, C. G., & van Moorsel, A. P. A. (2020, January 27). The relationship between trust in AI and trustworthy machine learning technologies. *FAT* '20: Proceedings of the 2020 Conference on Fairness, Accountability, and Transparency*, 272 - 283. https://doi.org/10.1145/3351095.3372834
United Nations. (2024, August 20). *Mapping report: human rights and new and emerging digital technologies* [Report of the Office of the United Nations High Commissioner for Human Rights]. United nations. https://www.ohchr.org/en/documents/thematic-reports/ahrc5645-mapping-report-human-rights-and-new-and-emerging-digital
Vagia, M., Transeth, A. A., & Fjerdingen, S. A. (2016, March). A literature review on the levels of automation during the years. What are the different taxonomies that have been proposed? *Applied Ergonomics*, *53*(A), 190-202. https://doi.org/10.1016/j.apergo.2015.09.013
van Breda, L., & Barry, T. (Eds.). (2020). *Human-Autonomy Teaming: Supporting Dynamically Adjustable Collaboration*. NATO / OTAN - North Atlantic Treaty Organization. https://apps.dtic.mil/sti/citations/AD1183655
Wang, L., Ma, C., Feng, X., Zhang, Z., Yang, H., Zhang, J., Chen, Z., Tang, J., Chen, X., Lin, Y., Zhao, W. X., Wei, Z., & Wen, J. (2024). A survey on large language model based autonomous agents. *Frontiers of Computer Science*, *18*(186345). https://doi.org/10.1007/s11704-024-40231-1
Wang, S., & Summers, R. M. (2012, July). Machine learning and radiology. *Medical Image Analysis*, *16*(5), 933-951. https://doi.org/10.1016/j.media.2012.02.005
Wei, J., Tay, Y., Bommasani, R., Raffel, C., Zoph, B., Borgeaud, S., Yogatama, D., Bosma, M., Zhou, D., Metzler, D., Chi, E. H., Hashimoto, T., Vinyals, O., Liang, P., Dean, J., & Fedus, W. (2022, August 31). Emergent Abilities of Large Language Models. *Transactions on Machine Learning Research*. https://openreview.net/forum?id=yzkSU5zdwD
Wei, J., Wang, X., Schuurmans, D., Bosma, M., Ichter, B., Xia, F., Chi, E. H., Le, Q. V., & Zhou, D. (2023). *Chain-of-Thought Prompting Elicits Reasoning in Large Language Models*. ArXiv. https://doi.org/10.48550/arXiv.2201.11903
Wilson, B., Natali, C., Roach, M. *et al.* (2025). Dimensions of Human-Machine Combination: Prompting the Development of Deployable Intelligent Decision Systems for Situated Clinical Contexts. *Comput Supported Coop Work* **34**, 425–481. https://doi.org/10.1007/s10606-025-09514-4
Winfield, A. F. T., & Jirotka, M. (2018, October 15). Ethical governance is essential to building trust in robotics and artificial intelligence systems. *Philosophical Transactions of the Royal Society A*, *376*(2133). https://doi.org/10.1098/rsta.2018.0085
Xi, Z., Chen, W., Guo, X., He, W., Ding, Y., Hong, B., Zhang, M., Wang, J., Jin, S., Zhou, E., Zheng, R., Fan, X., Wang, X., Xiong, L., Zhou, Y., Wang, W., Jiang, C., Zou, Y., Liu, X., … Gui, T. (2025, January 17). The rise and potential of large language model based agents: a survey. *Science China Information Sciences*, *68*(121101). https://doi.org/10.1007/s11432-024-4222-0
Yang, S., Nachum, O., Du, Y., Wei, J., Abbeel, P., & Schuurmans, D. (2023). *Foundation Models for Decision Making: Problems, Methods, and Opportunities*. arXiv. https://doi.org/10.48550/arXiv.2303.04129
Yelne, S., Chaudhary, M., Dod, K., Sayyad, A., & Sharma, R. (2023, November). Harnessing the Power of AI: A Comprehensive Review of Its Impact and Challenges in Nursing Science and Healthcare. *Cureus*, 22;15(11) : e49252. https://doi.org/10.7759/cureus.49252

# Appendix: Sectorial Relevance of the HAIG Approach for Analysing Trust Dynamics

Several implementation domains would particularly benefit from a continua approach to human-AI relationships rather than strict categorical frameworks (see Table 3). These sectors share characteristics that make the nuanced perspective of dimensions and continua especially valuable. From public services balancing democratic accountability with efficient delivery, to healthcare systems navigating graduated clinical authority, to financial services adapting to dynamic risk thresholds, each domain illustrates how AI governance exists along evolving spectrums rather than in fixed states.

The HAIG approach particularly valuable for sectors/applications with: high consequence decisions with varying stakes across contexts, professional judgement requirements that cannot be fully formalised, complex regulatory environments with proportionality requirements, strong ethical frameworks that already recognise degrees of responsibility, and risk profiles that vary across specific instances rather than fitting neat categories.

In these domains, the HAIG approach captures the nuanced reality of how human-AI relationships actually function while providing more precise governance guidance than categorical frameworks alone could offer. These sector-specific applications provide valuable comparative examples, demonstrating that the multidimensional HAIG framework has broad applicability across domains with similar characteristics.

| **Table 3:** Sectorial Relevance of the HAIG Approach for Analysing Trust Dynamics | |
|---|---|
| **Public Sector & Government Services** | • **Policy implementation spectrum**: Government services span from purely administrative functions to complex policy judgments requiring different degrees of human oversight<br>• **Stakeholder accountability**: Different public functions have varying levels of direct impact on citizens, requiring proportional human involvement<br>• **Democratic legitimacy considerations**: Different governance functions require different balances of efficiency and direct accountability<br>• **Cross-agency collaboration**: Public sector AI spans jurisdictional boundaries with varying authority distributions<br>• **Transparency requirements**: Different public functions have varying transparency needs that align with different points on the continuum |
| **Healthcare & Clinical Decision Support** | • **Graduated clinical authority**: Medical AI spans from simple reference tools to diagnostic partners with varying degrees of authority<br>• **Patient-specific risk levels**: Appropriate AI autonomy varies based on patient risk factors rather than fitting neat categories<br>• **Specialty-specific requirements**: Different medical specialties require different human-AI relationships<br>• **Professional ethics considerations**: Medical ethics frameworks address degrees of responsibility rather than binary states |
| **Financial Services & Risk Management** | • **Graduated risk thresholds**: Different transaction values and risk profiles require different levels of human oversight<br>• **Client-specific automation levels**: Appropriate automation levels vary based on client sophistication<br>• **Market condition adaptivity**: Human-AI relationships need to adjust with changing market volatility<br>• **Regulatory proportionality**: Financial regulations increasingly apply proportionate controls based on risk levels |
| **Legal Services & Judicial Processes** | • **Jurisdictional variations**: Different legal systems permit different levels of automation<br>• **Case complexity spectrum**: Simple cases may permit more automation than complex ones<br>• **Stakes-based oversight**: Higher-stakes matters require different human-AI relationships than routine matters<br>• **Legal authority gradations**: Legal authority exists along a spectrum rather than in discrete categories |
| **Education & Learning Assessment** | • **Developmental appropriateness:** Different student ages and stages require different levels of human involvement<br>• **Subject-specific variations:** Some subjects permit more automation than others<br>• **Pedagogical diversity:** Different teaching philosophies align with different points on the human-AI continuum<br>• **Stakes-based considerations:** High-stakes assessment requires different human-AI relationships than practice activities |
| **Critical Infrastructure Management** | • **Graduated criticality levels:** Different systems and components have varying criticality requiring different oversight<br>• **Temporal variations:** Appropriate automation levels change during normal operations versus emergencies<br>• **Physical-digital integration:** Systems span from purely advisory to having direct control of physical components<br>• **Cascading consequence management:** Oversight needs scale with potential for cascading failures |
| **Research & Scientific Discovery** | • **Methodological diversity:** Different research methods permit different levels of automation<br>• **Discovery-validation spectrum:** Early discovery phases may permit more AI agency than validation phases<br>• **Field-specific standards:** Different disciplines have different standards for evidence and verification<br>• **Risk-reward calculations:** Higher-risk research domains require different human-AI relationships |